\documentclass[letterpaper]{article} 
\usepackage{aaai2026}  
\usepackage{times}  
\usepackage{helvet}  
\usepackage{courier}  
\usepackage[hyphens]{url}  
\usepackage{graphicx} 
\usepackage{natbib}  
\usepackage{caption} 
\usepackage{algorithm}
\usepackage{algorithmic}

\usepackage{newfloat}
\usepackage{listings}
\DeclareCaptionStyle{ruled}{labelfont=normalfont,labelsep=colon,strut=off} 
\floatstyle{ruled}
\newfloat{listing}{tb}{lst}{}
\floatname{listing}{Listing}
\usepackage{booktabs}       
\usepackage{amsfonts, amsmath, amssymb, bm} 
\usepackage{nicefrac}       
\usepackage{microtype}      
\usepackage{graphicx}       
\usepackage{multirow, makecell, array, colortbl, tabularx} 
\usepackage{bbding}         
\usepackage{newfloat}       
\usepackage{listings}       
\usepackage{xpatch}         
\usepackage{afterpage}

\title{RMLer: Synthesizing Novel Objects across Diverse Categories \\ via Reinforcement Mixing Learning}

\author{
    Jun Li\textsuperscript{\rm 1}\thanks{Corresponding authors.},
    Zikun Chen\textsuperscript{\rm 1},
    Haibo Chen\textsuperscript{\rm 1}\footnotemark[1],
    Shuo Chen\textsuperscript{\rm 2},
    Jian Yang\textsuperscript{\rm 2}
}
\affiliations{
    \textsuperscript{\rm 1}School of Computer Science and Engineering, Nanjing University of Science and Technology, China\\
    \textsuperscript{\rm 2}School of Intelligence Science and Technology, Nanjing University, China\\
    \texttt{\{junli, zikunchen, hbchen\}@njust.edu.cn,\ \{shuo.chen, csjyang\}@nju.edu.cn}
}

\begin{document}

\maketitle



\begin{abstract}
Novel object synthesis by integrating distinct textual concepts from diverse categories remains a significant challenge in Text-to-Image (T2I) generation. Existing methods often suffer from insufficient concept mixing, lack of rigorous evaluation, and suboptimal outputs—manifesting as conceptual imbalance, superficial combinations, or mere juxtapositions. To address these limitations, we propose \textbf{Reinforcement Mixing Learning (RMLer)}, a framework that formulates cross-category concept fusion as a reinforcement learning problem: mixed features serve as states, mixing strategies as actions, and visual outcomes as rewards. Specifically, we design an MLP-policy network to predict dynamic coefficients for blending cross-category text embeddings. We further introduce visual rewards based on (1) semantic similarity and (2) compositional balance between the fused object and its constituent concepts, optimizing the policy via proximal policy optimization. At inference, a selection strategy leverages these rewards to curate the highest-quality fused objects. Extensive experiments demonstrate RMLer's superiority in synthesizing coherent, high-fidelity objects from diverse categories, outperforming existing methods. Our work provides a robust framework for generating novel visual concepts, with promising applications in film, gaming, and design.
\end{abstract}

\begin{figure}[t]
\centering
\includegraphics[width=\linewidth]{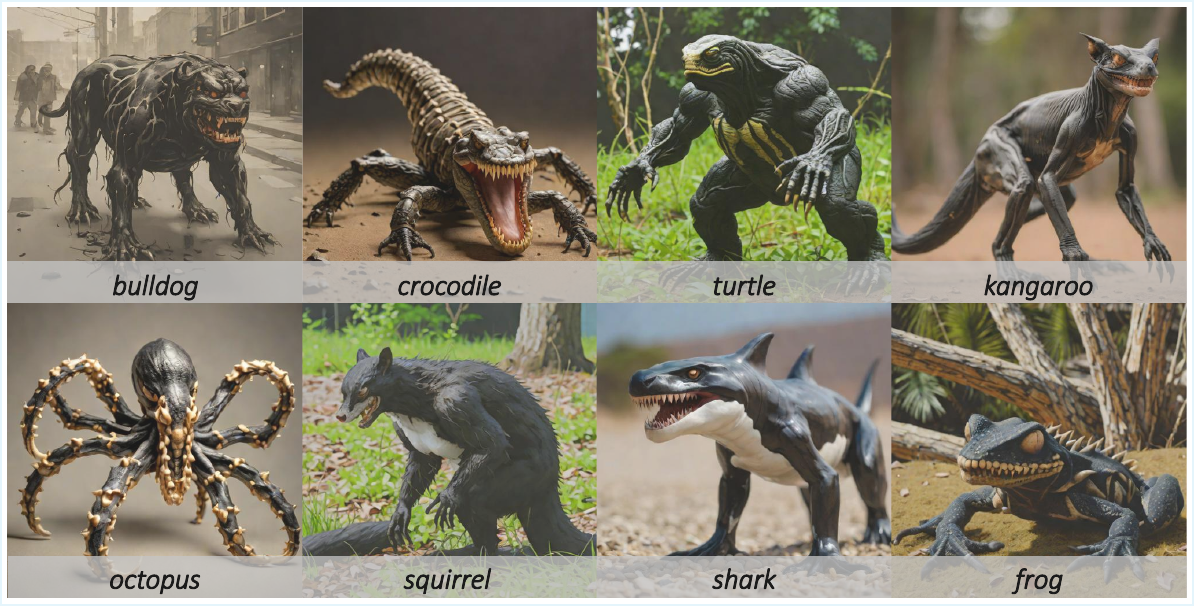}
\caption{
We propose a simple yet effective reinforcement mixing learning approach for generating novel object images by fusing distinct categories. 
For instance, our method seamlessly combines the \textit{Venom} character with diverse animal categories—such as 
\textit{bulldog}, \textit{crocodile}, \textit{turtle}, \textit{kangaroo}, and \textit{frog}—effectively blending their features to demonstrate its versatility.
}
\label{fig:fig1}
\end{figure}


\section{Introduction}
\label{sec:introduction}

The rise of large-scale Text-to-Image (T2I) synthesis, driven primarily by advances in diffusion models~\cite{rombach2022high, saharia2022photorealistic, podell2023sdxl, esser2024scaling}, has revolutionized digital content creation. These systems now support a wide range of applications, from artistic design~\cite{wang2024diffusion, horvath2024ai, paananen2024using, montenegro2024integrative, jin2025compose} and virtual reality~\cite{yin2024text2vrscene,behravan2025generative} to film production and game development~\cite{zhou2024eyes, sun2024text2ac}. Recent improvements in photorealistic fidelity~\cite{openai2024dalle3, chen2024vividdreamer} and output diversity~\cite{Bau2023editing} have pushed the boundaries of what these systems can achieve. The next frontier lies in enhancing compositional reasoning and fine-grained control~\cite{zhang2023controlnet, meng2022sdedit}, particularly in synthesizing novel objects by combining features from multiple concepts across different categories.


\begin{figure*}[t]
  \centering
  \includegraphics[width=1\linewidth]{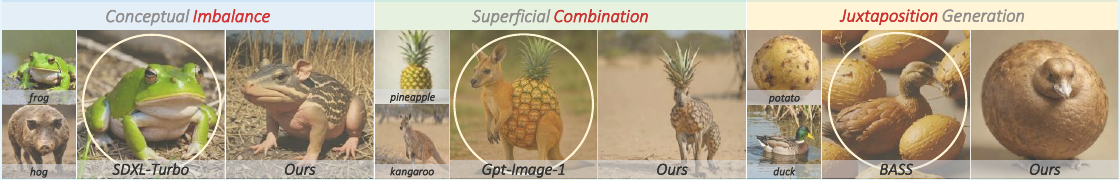}
  \caption{Failures in concept fusion by existing methods. \textbf{Left} (SDXL-Turbo~\cite{podell2023sdxl}): Severe imbalance (e.g., \textit{frog} + \textit{hog} $\rightarrow$ dominant frog). \textbf{Middle} (GPT-Image-1): Superficial combination (e.g., \textit{pineapple} + \textit{kangaroo}). \textbf{Right} (BASS~\cite{li2024tp2o}): Simple juxtaposition (e.g., \textit{owl} + \textit{snail}). Our approach (rightmost) aims for more balanced and coherent fusions.}
  \label{fig:motivation_examples}
\end{figure*}

Current T2I diffusion models employ two primary approaches to fuse multiple distinct textual concepts into a single, coherent object: general-purpose foundational models (e.g., SDXL-Turbo~\cite{podell2023sdxl}, DALL·E 3~\cite{openai2024dalle3}, Flux~\cite{flux2024}, GPT-Image-1~\cite{openai_gptimage1}) and specialized fusion techniques (e.g., BASS~\cite{li2024tp2o}, ConceptLab~\cite{Richardson2024conceptlab}). Despite their capabilities, these methods exhibit three key limitations: (1)
\textbf{Conceptual Imbalance}–-The generated image predominantly represents one object category, significantly overshadowing the other (left, Fig.~\ref{fig:motivation_examples}). This bias stems from imbalanced prompt features, allowing one concept to dominate the composition.
(2) \textbf{Superficial Combination}–-The two concepts are merely overlapped without meaningful integration (middle, Fig.~\ref{fig:motivation_examples}). Due to imbalanced local prompt features, the model exhibits a bias toward certain concepts in different spatial regions, disrupting coherent integration. (3) \textbf{Juxtaposition Generation}–-The objects are placed separately in the image rather than being fused (right, Fig.~\ref{fig:motivation_examples}). Without precise spatial control, the model generates multiple objects rather than a unified composition. Fundamentally, these issues arise from insufficient mixing and control over the characteristic features of the two categories.


To address these limitations, we propose \textbf{Reinforcement Mixing Learning (RMLer)}, a novel framework that formulates cross-category concept fusion as a reinforcement learning (RL) problem. Given two category labels, we first extract their text embeddings or features using simple prompt: \textit{A photo of $<$category label$>$}, and define element-wise interpolation between these embeddings as a mixing strategy. The core idea of RMLer is to treat mixed text features as states, mixing strategies as actions, and the resulting visual outputs as rewards. 
Specifically, we design an MLP-policy network to predict dynamic interpolation coefficients for blending cross-category text embeddings. To guide learning, we introduce visual rewards that measure both semantic similarity and compositional balance between the fused object and its constituent concepts. These rewards ensure that the mixed features effectively integrate (local) prompt features, mitigating conceptual imbalance and superficial combination. Additionally, we leverage a foreground segmentation model to isolate objects in generated images, avoiding unintended juxtapositions. 
The policy network is optimized via Proximal Policy Optimization (PPO)~\cite{schulman2017proximal}. During inference, a principled post-selection mechanism—guided by metrics aligned with our reward functions—refines the outputs to select the most compelling fused objects. 
The key strength of RMLer lies in its ability to learn adaptive mixing strategies through direct optimization of complex fusion objectives, enabling sophisticated embedding manipulation. 
Extensive experiments demonstrate RMLer’s effectiveness, showing that it generates novel fused objects that standard text-to-image baselines struggle to produce (see Fig.~\ref{fig:res1}). Overall, our main contributions are as follows. 
\begin{itemize}
    \item 1) We propose Reinforcement Mixing Learning, a framework that learns an adaptive policy to dynamically manipulate text embeddings across diverse categories for generating novel objects. To the best of our knowledge, this is the first work to effectively formulate cross-category fusion as a reinforcement learning problem.
    \item 2) Extensive experiments (Fig.~\ref{fig:motivation_examples}, Table~\ref{tab:quant_cmp_splitcol_extended}) demonstrate that RMLer synthesizes harmoniously fused objects from disparate categories, outperforming standard T2I techniques in quality, coherence, and compositional fidelity.
\end{itemize}
 
\begin{figure*}[t]
    \centering
    \includegraphics[width=0.92\textwidth]{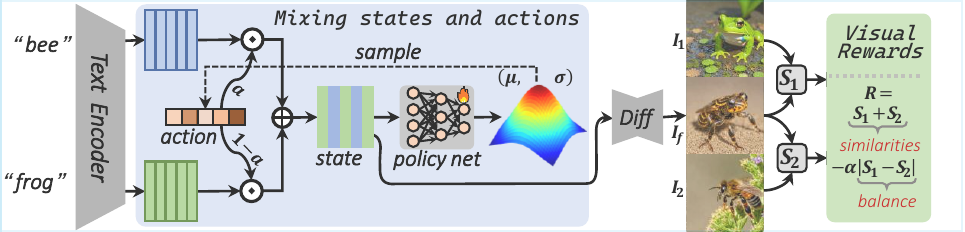} 
    \caption{Pipeline of our Reinforcement Mixing Learning (RMLer). Given CLIP embeddings for two concepts ($\mathbf{e}_1, \mathbf{e}_2$) extracted from labels ($c_1, c_2$), our policy network $\pi_{\theta}$ generates an action vector $\mathbf{a}$ that mixs $\mathbf{e}_1$ and $\mathbf{e}_2$ into a fused embedding $\mathbf{e}_f$. This embedding conditions a diffusion model $\mathcal{G}$ to synthesize the image $I_f$. A visual reward $R$, computed from CLIP similarity and balance between $I_f$ and references $I_1$ and $I_2$ generated by $\mathbf{e}_1$ and $\mathbf{e}_2$ respectively, guides the PPO algorithm to update $\pi_{\theta}$. } 
    \label{fig:method_overview} 
\end{figure*}

\section{Related Work}
\label{sec:related_work}

Text-to-image (T2I) synthesis has advanced rapidly, but controllable and coherent fusion of multiple concepts remains challenging. We review related work in three areas: T2I synthesis, alignment of diffusion models, and object fusion.

\textbf{Text-to-Image Synthesis.}  
Diffusion models have significantly advanced T2I synthesis in fidelity and diversity~\cite{rombach2022high, saharia2022photorealistic, podell2023sdxl, zhang2023controlnet, Zhang2023inversion, Gu2022vqdiffusion, gong2024text2avatar}. Recent architectures like MMDiT~\cite{esser2024scaling} further improve multi-entity and stylistic generation. However, these models are primarily designed for holistic scene synthesis from a single prompt and often fail when fusing distinct concepts into a coherent entity. Challenges such as attribute leakage~\cite{roy2019mitigating} and semantic imbalance~\cite{ma2022delving} arise, especially for out-of-distribution combinations~\cite{madan2022and}. In contrast, RMLer presents a policy-driven control over input conditioning, learning to optimally merge concept embeddings for improved compositional fusion.

\textbf{Alignment of Diffusion Models.}  
To enhance controllability in diffusion models, recent works leverage reinforcement learning from human feedback (RLHF)~\cite{liu2024efficient, liu2024improving}, widely adopted in large language model alignment~\cite{ouyang2022training, bai2022training}. Reward models~\cite{schuhmann2022laion, xu2023imagereward, kirstain2023pick, wu2023hpsv2} have enabled learning-based guidance in image generation. Building on this, DDPO~\cite{black2023training}, DPOK~\cite{fan2023dpok}, DiffusionDPO~\cite{wallace2024diffusion}, and others~\cite{clark2023directly, prabhudesai2023aligning} formulate diffusion sampling as an MDP and apply policy gradients or reward backpropagation for alignment. While effective at attribute control, these methods typically modify the backbone. In contrast, RMLer introduces a lightweight policy over conditioning embeddings, enabling fine-grained fusion without altering the diffusion network.

\textbf{Object Fusion.}  
There has been increasing interest in generating fused images \cite{liew2022magicmix, yi2024diff} from multiple concepts, a task that holds great potential for creative applications such as digital art and design. ConceptLab~\cite{Richardson2024conceptlab} employs diffusion models to synthesize unique visual concepts but its optimization-based approach is computationally expensive and often struggles to semantically integrate real-world concepts. BASS~\cite{li2024tp2o} introduces a more controllable framework for concept fusion by learning balance-aware token swapping. However, the swapped regions can sometimes lead to non-meaningful or visually chaotic results. In contrast, our RMLer framework offers a more efficient and adaptive solution for concept fusion. By learning a policy to directly manipulate embeddings, RMLer enables faster generation of semantically coherent and well-balanced fused images.

\section{Preliminaries}
\label{sec:preliminaries} 


\textbf{Markov Decision Process (MDP)} \cite{garcia2013markov} provides a mathematical framework for modeling decision-making under uncertainty. An MDP is defined by a tuple $(\mathcal{S}, \mathcal{A}, P, R, \rho_0)$, where $\mathcal{S}$ and $\mathcal{A}$ denote the state and action spaces, $P$ is the state transition probability, $R$ is the reward function, and $\rho_0$ is the initial state distribution. At each step, an agent selects an action $\mathbf{a}_t \sim \pi(\mathbf{a}_t|\mathbf{s}_t)$, receives a reward $R(\mathbf{s}_t, \mathbf{a}_t)$, and transitions to a new state $\mathbf{s}_{t+1}$. The goal of reinforcement learning is to find a policy $\pi^*$ that maximizes the expected cumulative reward:
\begin{equation}
    \mathcal{J}_\text{RL}(\pi) = \mathbb{E}_{\tau \sim p(\tau \mid \pi)} \left[ \sum\nolimits_{t=0}^T R(\mathbf{a}_t, \mathbf{s}_t) \right],
\label{eq:rl_objective_finite_horizon}
\end{equation}
where $\tau$ is the trajectory generated by a ploicy $\pi$ over $T$.

\textbf{Denoising Diffusion Policy Optimization (DDPO)}~\cite{black2023training} reformulates the iterative denoising process of diffusion models as a multi-step MDP to enable fine-tuning via reinforcement learning. Each denoising step is treated as an action, and the policy $\pi$ corresponds to the reverse diffusion kernel $p_\theta(\mathbf{x}_{t-1} \mid \mathbf{x}_t, \mathbf{c})$, conditioned on time $t$ and context $\mathbf{c}$. The MDP components are defined as:

\begin{align*}
   & \mathbf{s}_t \triangleq (\mathbf{c}, t, \mathbf{x}_t), 
   \ \ \ \ \ \ \
   \pi(\mathbf{a}_t \mid \mathbf{s}_t) \triangleq 
   p_\theta(\mathbf{x}_{t-1} \mid \mathbf{x}_t, \mathbf{c}), \\[4pt]
   & \mathbf{a}_t \triangleq \mathbf{x}_{t-1},  
   \ \ \ \ \ \ \ \ \ \ \ \ \ 
   \rho_0(\mathbf{s}_T) \triangleq 
   (p(\mathbf{c}), \delta_T, \mathcal{N}(\mathbf{0}, \mathbf{I})), \\[4pt]
   & P(\mathbf{s}_{t-1} \mid \mathbf{s}_t, \mathbf{a}_t) 
   \triangleq 
   (\delta_{\mathbf{c}}, \delta_{t-1}, \delta_{\mathbf{x}_{t-1}}), 
   \\
   & R(\mathbf{s}_t, \mathbf{a}_t) \triangleq
        \begin{cases}
            r(\mathbf{x}_0, \mathbf{c}) & \text{if } t = 0, \\
            0 & \text{otherwise},
        \end{cases}
\end{align*}

where $\delta_y$ denotes the Dirac delta function.

\section{Methodology}
\label{sec:methodology}
In this section, we present a Reinforcement Mixing Learning (RMLer) framework for multi-concept fusion in Fig. \ref{fig:method_overview}. Our approach consists of three key components. \textit{Problem Formulation:} We formulate multi-concept fusion as a reinforcement learning (RL) task.
\textit{Visual Reward Function:} We introduce a reward function based on visual similarity and balance, ensuring high-quality and harmonious outputs. \textit{Two-Stage Sampling Strategy:} To enhance efficiency, we propose a two-stage sampling method that selects the most representative fused object from candidate generations.

\subsection{Problem Formulation of RMLer} 
\label{subsec:rmler_multistep_mdp}


Cross-category concept fusion (CCF) is a challenging task that combines two distinct textual concepts, $c_1$ and $c_2$ into a single novel and coherent object image $I_f$. Our RMLer formulates this task as a multi-step Markov Decision Process (MDP), shown in Figure~\ref{fig:method_overview}. Before detailing our method, we first formally define the CCF task.

\textbf{The CCF Task.} Given two distinct category labels \(c_1\) and \(c_2\), we construct simple text prompts: $p_1:\text{\textit{A photo of}} <c_1>$ and $ p_2:\text{\textit{A photo of}} <c_2>$. These prompts are fed into the T2I diffusion model to generate their corresponding original images: $I_1\sim \mathcal{G}(\mathbf{e}_1,\epsilon)\in \mathbb{R}^{H\times W}$ and $I_2\sim\mathcal{G}(\mathbf{e}_2,\epsilon)\in \mathbb{R}^{H\times W}$, where $\mathbf{e}_1=\mathcal{E}(p_1)\in \mathbb{R}^{h\times w}$, $\mathbf{e}_2=\mathcal{E}(p_2)\in \mathbb{R}^{h\times w}$ and \(\epsilon\) is a sampling noise. The CCF task involves fusing \(\mathbf{e}_1\) and \(\mathbf{e}_2\) into a mixing text embedding \(\mathbf{e}_f\), which is then used to generate a novel and coherent fused image, $I_f=\mathcal{G}(\mathbf{e}_f)\in \mathbb{R}^{H\times W}$. In our implementation, we use a pretrained Stable Diffusion model \cite{podell2023sdxl} as our baseline, where $\mathcal{E}(\cdot)$ denotes the text encoder and $\mathcal{G}(\cdot)$ represents the diffusion-based generator. Our framework is model-agnostic and can be adapted to other diffusion models.

\textbf{CCF as a multi-step MDP.} We formulate the CCF task as a multi-step (MDP). In each fusion episode, consisting of $T$ steps ($t = 0, \dots, T{-}1$), the agent interacts with the environment as follows:


\begin{itemize}
    \item \textbf{State $\mathbf{s}_t$:}  the current fused embedding $\mathbf{e}^{(t)}_f\in \mathbb{R}^{h\times w}$.
    \item \textbf{Action $\mathbf{a}_t$:} the column-wise interpolation coefficient $\mathbf{a}_t \in \mathbb{R}^w$. The initial state $\mathbf{s}_0$ is computed as the average of the source embeddings, \(\mathbf{s}_0 = \mathbf{e}^{(0)}_f = \frac{1}{2}(\mathbf{e}_1 + \mathbf{e}_2)\).
    \item \textbf{Policy $\pi_{\theta}(\mathbf{a}_t|\mathbf{s}_t)$:} a stochastic policy parameterized by an MLP with weights $\theta$,which outputs a distribution over possible actions given the current state  $\mathbf{s}_t$. An action is sampled as \(\mathbf{a}_t \sim \pi_{\theta}(\cdot \mid \mathbf{s}_t)\).
    \item \textbf{Transition:} Updates the state $\mathbf{s}_t$ to the next state $\mathbf{s}_{t+1}$ via a fusion function $f_{\text{fuse}}(\cdot)$:
    \begin{align}
         \mathbf{e}^{(t+1)}_f =& f_{\text{fuse}}(\mathbf{a}_t, \mathbf{e}_1, \mathbf{e}_2) \nonumber  \\
        =&\mathbf{e}_1 \times \text{diag}(\mathbf{a}_t) + \mathbf{e}_2 \times \text{diag}(1-\mathbf{a}_t),
        \label{eq:general_fusion}
    \end{align}
    where $\text{diag}(\mathbf{a}_t)$ denotes converting the vector \(\mathbf{a}_t\) into a diagonal matrix.
     \item \textbf{Reward:} an evaluation score \(R_{t+1} = r(I^{(t+1)}_f, c_1, c_2)\), where \(I^{(t+1)}_f \sim \mathcal{G}(\mathbf{e}^{(t+1)}_f, \epsilon)\). 
\end{itemize}

\textbf{Formal MDP at timestep $t$:}
\begin{align}
    \mathbf{s}_t &\triangleq \mathbf{e}^{(t)}_f,  &
    \pi_\theta(\mathbf{a}_t \mid \mathbf{s}_t) &\triangleq P(\mathbf{a}_t \mid \mathbf{s}_t; \theta),  \nonumber  \\
    \mathbf{a}_t &\sim \pi_\theta(\cdot \mid \mathbf{s}_t),  &
    \mathbf{s}_{t+1} &\triangleq f_{\text{fuse}}(\mathbf{a}_t, \mathbf{e}_1, \mathbf{e}_2) ,
    \label{eq:rmler_multistep_mdp_def} \\
    I^{(t+1)}_f &\sim \mathcal{G}(\mathbf{s}_{t+1}, \epsilon)  &
    R_{t+1}(\mathbf{s}_t, \mathbf{a}_t) &\triangleq r(I^{(t+1)}_f, c_1, c_2).  \nonumber 
\end{align}

The RMLer objective is to learn $\pi_\theta$ that maximizes the quality of the best fused result encountered within an $T$-step trajectory $\{\mathbf{s}_0, \mathbf{a}_0, \dots, \mathbf{s}_T\}$. While the process yields a sequence of images with rewards $\{R_1, \dots, R_T\}$, our primary goal is to maximize the highest single-step reward, i.e., $\max_t R_t$, where we use $\sum_{t=1}^T \gamma_t R_t$ with $\gamma_t = 1$ (no discounting) to aggregate rewards. To guide policy learning, we retain intermediate rewards at each step and optimize $\pi_\theta$ using Proximal Policy Optimization (PPO) \cite{schulman2017proximal}, with the surrogate objective: 
\begin{align}
\mathcal{L}^{\text{PPO}}(\theta) = & \mathbb{E}_{(s_t, a_t) \sim \pi_{\theta_{\text{old}}}} \left[
    - \min \left(k_t(\theta) \cdot R(s_t, a_t), \right. \right. \nonumber \\
      &  \left. \left.
        \text{clip}\left(k_t(\theta),\, 1 - \xi, 1 + \xi
        \right) \cdot R(s_t, a_t)
    \right)
\right],
\label{eq:ppo}
\end{align}
where $k_t(\theta)= \frac{\pi_\theta(a_t \mid s_t)}{\pi_{\theta_{\text{old}}}(a_t \mid s_t)}$ is a probability ratio, and $\xi=0.2$ is a hyperparameter. Unlike standard PPO, our formulation in Eq. \eqref{eq:ppo} eliminates the critic network, relying solely on policy optimization. Table \ref{tab:ppo_ablation} presents preliminary experiments showing that standard PPO suffers from performance degradation due to unstable training dynamics.  To address this issue, we introduce an intrinsic visual reward mechanism in the following subsection.

\begin{table}[t]
\centering
\small
\caption{Comparing PPO and our RMLer in CangJie-200.}
\label{tab:ppo_ablation}
\setlength{\tabcolsep}{3pt}
\renewcommand{\arraystretch}{1.1}
\scalebox{0.87}{
\begin{tabular}{c|cc|c|cc}
\Xhline{1.2pt} 
-- &  \textbf{RMLer}  &   PPO   &-- &  \textbf{RMLer}  &   PPO   \\
\hline  
HPSv2 \(\uparrow\) & \textbf{0.2774} & 0.2746  & VQAScore \(\uparrow\) & \textbf{0.4287} &  0.4155  \\
\Xhline{1.2pt} 
\end{tabular} }
\end{table}





\begin{figure*}[t]
    \centering
    \includegraphics[width=0.85\textwidth]{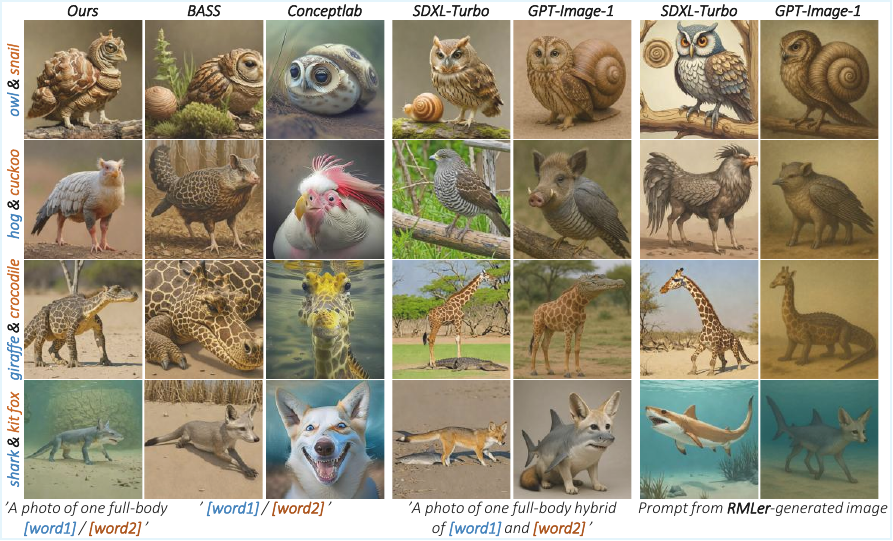}
    \caption{
    Comparisons with different methods on \textbf{ImageNet-200}. The complex prompts are created from RMLer-generated image using GPT-4o. For instance, \textit{A hybrid creature combining an owl and a snail, with an owl-like head, sharp eyes and a curved beak, and a body covered by a spiral shell texture, standing on a wooden branch with bird-like legs and claws.}
    }
    \label{fig:res1}
\end{figure*}

\subsection{Visual Reward Function}
\label{subsec:reward_function}
The reward \( r(I_f, c_1, c_2) \) plays a key role in evaluating our method for the CCF task. We present a visual reward function that  based on CLIP similarity and balance between $I_f$ and reference exemplars $I_1$ and $I_2$, which are generated by $\mathbf{e}_1(c_1)$ and $\mathbf{e}_2(c_2)$, respectively. 

Specifically, we first extract foreground segments---\(I_{f\text{seg}}\), \(I_{1\text{seg}}\), and \(I_{2\text{seg}}\)---from \(I_f\), $I_1$ and $I_2$ using a foreground segmentation model \cite{oquab2024dinov2}. We then compute their CLIP image embeddings \(\mathbf{f}_{I_{f\text{seg}}}\),  \(\mathbf{f}_{I_1}\) and \(\mathbf{f}_{I_2}\) via a pretrained CLIP image encoder \(E_{\text{CLIP-I}}\) \cite{radford2021learning}. The \textbf{visual fusion reward} is defined as: 
\begin{equation}
R = (S_1 + S_2) - \alpha \cdot |S_1 - S_2|,
\label{eq:reward_function_final}
\end{equation}
where $S_1 = \text{sim}(\mathbf{f}_{I_{f\text{seg}}}, I_{1\text{seg}})$ and $S_2 = \text{sim}(\mathbf{f}_{I_{f\text{seg}}}, I_{2\text{seg}})$ denote the cosine similarities between the generated image and the two concept exemplars. The first two terms ensure that the fused image $I_f$ maintains maximum similarity with both $I_1$ and $I_2$, indicating that $I_f$ retains more characteristics from the distinct categories $c_1$ and $c_2$. The last term promotes balanced alignment with both concepts. The scale factor \(\alpha > 0\) mitigates excessive dominance of one concept over the other. 

This formulation encourages the RMLer policy \(\pi_\theta\) to explore embedding manipulations that produce visually coherent and semantically balanced fusion results. Empirically, we find that image-based CLIP similarity offers stronger guidance than text-based reward signals, a finding further supported by our ablation studies (\textbf{\textit{refer to Appx.~A}}).

\subsection{Two-Stage Sampling Strategy}
\label{subsec:image_selection}
After learning the policy $\pi_{\theta^*}$  (typically the best-performing checkpoint),  the stochasticity of both policy sampling and diffusion generation leads to variability in inference outputs. To identify representative examples that reliably capture the capabilities of our RMLer framework—particularly for qualitative evaluation and visualization—we introduce a principled two-stage selection strategy.

\textbf{Candidates Selected via Fusion Criteria.}
In the first stage, we filter a larger pool of generated images $\mathcal{I}_f=\{I_f\}$ to obtain a candidate set that meets two core criteria: \textit{Concept Presence} means that the fused image must clearly exhibit the semantic attributes of the input categories; and \textit{Fusion Balance} shows tht the composition should harmoniously integrate all relevant elements. An image $I_f$ is retained as a candidate only if it satisfies both conditions:
\begin{enumerate}
    \item \textit{Dual Concept Presence:} $S_1 > \tau_{\text{presence}}$ and $S_2 > \tau_{\text{presence}}$.
    \item \textit{Fusion Balance:} $|S_1 - S_2| < \tau_{\text{balance}}$.
\end{enumerate}
where $\tau_{\text{presence}}$ and $\tau_{\text{balance}}$ are empirically set to $0.63$ and $0.05$, respectively.
Therefore, we have a candidate set:
\begin{align}
\mathcal{I}_{\text{can}}= \{  I_f\mid & \ S_1 > \tau_{\text{presence}} \ \& \ S_2 > \tau_{\text{presence}} \ \&  \nonumber \\ & \ |S_1 - S_2| < \tau_{\text{balance}}, I_f \in\mathcal{I}_f\}.
\label{eq:can}
\end{align}
\textbf{Top-$1$ Ranking.} From the set $\mathcal{I}_{\text{can}}$, we select the top-$1$ image with the highest total semantic alignment score, computed as the sum of its similarities to both source concepts:
\begin{equation}
I_f^* =\max_{I_f\in\mathcal{I}_{\text{can}}} S_1 + S_2,
\label{eq:pfs_for_selection}
\end{equation}
where the top-$1$ image with the highest score is selected as the final representative exemplar. Of course, the top-$K$ images can also be provided for user selection.

By decoupling fusion balance (enforced during candidate qualification) from concept preservation strength, our two-stage selection ensures that the chosen exemplars are both semantically balanced and highly representative. This approach simplifies scoring while avoiding redundancy.

\section{Experiments}

\subsection{Experimental Settings}
\label{sec:settings}

\textbf{Dataset.}
We evaluate on a benchmark of 400 diverse concept pairs. This includes \textbf{ImageNet-200}, a set of 200 manually curated pairs from the ImageNet~\cite{russakovsky2015imagenet} vocabulary, selected to maximize semantic and visual dissimilarity. In addition, we incorporate the \textbf{CangJie} dataset proposed in CreTok~\cite{feng2024redefining}, which contains 200 concept pairs designed to test compositional creativity in the TP2O~\cite{li2024tp2o} task.

\textbf{Details.} Our method is implemented based on the \textbf{SDXL-Turbo} model~\cite{podell2023sdxl} for efficient text-to-image generation. For semantic feature extraction—used in both reward computation and image selection—we employ the \textbf{CLIP ViT-H/14} model~\cite{radford2021learning}. Foreground segmentation is performed using the \textbf{RMBG-2.0} model~\cite{BiRefNet} to isolate salient content for evaluation. All generated and processed images are standardized to a resolution of $512 \times 512$ pixels. Experiments were conducted on a system equipped with two NVIDIA GeForce RTX 4090 GPUs.

\begin{figure*}[t]
    \centering
    \includegraphics[width=0.85\textwidth]{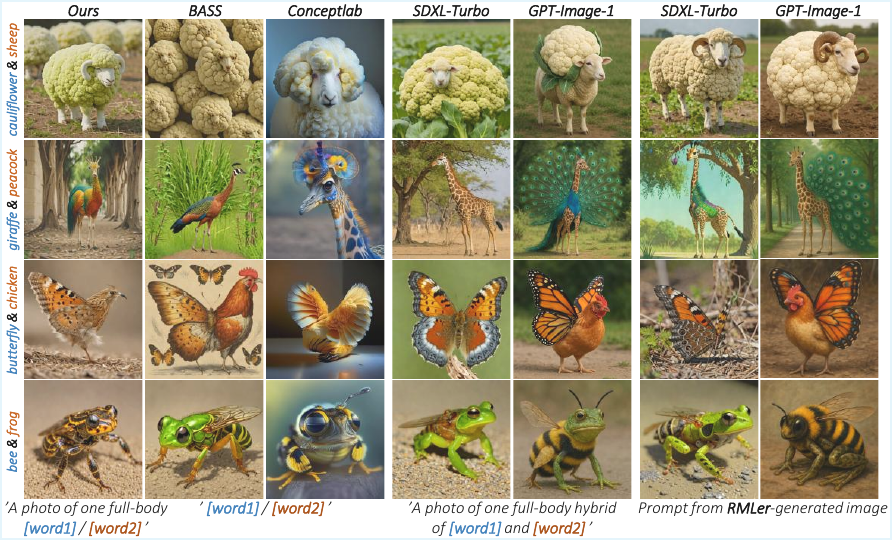} 
    \caption{ Comparison with different methods on the \textbf{CangJie-200}. The complex prompts are created from RMLer-generated image using GPT-4o. For instance, \textit{A hybrid creature combining a butterfly and a chicken, with a compact, feathery bird body, thin legs and claws, and large, vibrant butterfly wings extending from its back, standing on dry forest ground.} }
    \label{fig:res2}
\end{figure*}

\textbf{Evaluation Metrics.}
We evaluate the performance of our RMLer framework using a comprehensive set of automated metrics that assess both fusion quality and perceptual realism. Specifically, we report the following five metrics:

\begin{itemize}
    \item \textbf{Avg. Sim (I$\rightarrow$I / I$\rightarrow$T)}$\uparrow$: 
    Mean CLIP similarity between the generated image and either exemplar images (I$\rightarrow$I) or text prompts (I$\rightarrow$T), measuring overall concept alignment.    
    \item \textbf{Balance (I$\rightarrow$I / I$\rightarrow$T)}$\downarrow$: 
    Absolute difference between CLIP similarities to the two source concepts (images or texts); lower values indicate more balanced fusion.
    \item \textbf{Reward}$\uparrow$: Our reward score computes $(S_1 + S_2) - \alpha |S_1 - S_2|$ in Eq.~\ref{eq:reward_function_final}, balancing concept presence and symmetry.    
    \item \textbf{HPSv2}$\uparrow$ \cite{wu2023hpsv2} estimates human preference alignment for generated images, capturing overall aesthetic and alignment qualities.    
    \item \textbf{VQAScore}$\uparrow$ \cite{lin2024evaluating} evaluates visual-text alignment in complex compositional prompts.    
\end{itemize}
These metrics provide a comprehensive assessment across fusion accuracy, conceptual balance, and perceptual quality.

\begin{table*}[t]
\centering
\caption{Quantitative comparison on the ImageNet-200 and CangJie-200 benchmarks. For Avg. Sim and Balance, we report both image-to-image (I$\rightarrow$I) and image-to-text (I$\rightarrow$T) variants.}
\label{tab:quant_cmp_splitcol_extended}
\setlength{\tabcolsep}{3pt}
\renewcommand{\arraystretch}{1.1}
\scalebox{0.77}{
\begin{tabular}{l|
cc|cc|
cc|cc|
cc|cc|
cc}
\Xhline{1.2pt} 
\multirow{2}{*}{Model} & 
\multicolumn{2}{c|}{Avg. Sim (I$\rightarrow$I)$\uparrow$} & 
\multicolumn{2}{c|}{Avg. Sim (I$\rightarrow$T)$\uparrow$} & 
\multicolumn{2}{c|}{Balance (I$\rightarrow$I)$\downarrow$} & 
\multicolumn{2}{c|}{Balance (I$\rightarrow$T)$\downarrow$} & 
\multicolumn{2}{c|}{Reward$\uparrow$} & 
\multicolumn{2}{c|}{HPSv2$\uparrow$} & 
\multicolumn{2}{c}{VQAScore$\uparrow$} \\
& Img & CJ & Img & CJ & Img & CJ & Img & CJ & Img & CJ & Img & CJ & Img & CJ \\
\hline
\textbf{Our RMLer} 
& \underline{0.7324} & \underline{0.7193} 
& 0.2272 & \underline{0.2452}
& \textbf{0.0080} & \textbf{0.0070} 
& \textbf{0.0394} & \textbf{0.0364} 
& \textbf{1.4244} & \textbf{1.4034} 
& \underline{0.2737} & \textbf{0.2774} 
& \textbf{0.3301} & \textbf{0.4287} \\
BASS~\cite{li2024tp2o} 
& 0.7026 & 0.6595 
& 0.2223 & 0.2219
& 0.0918 & 0.1309 
& \underline{0.0659} & 0.0830 
& 0.9459 & 0.6640 
& \textbf{0.2756} &\underline{ 0.2750} 
& \underline{0.3055} & 0.3069 \\
ConceptLab~\cite{Richardson2024conceptlab} 
& 0.5991 & 0.6021 
& 0.2211 & 0.2434
& \underline{0.0908} & 0.1112 
& 0.0701 & 0.0662 
& 0.7440 & 0.6480 
& 0.2636 & 0.2714 
& 0.2671 & \underline{0.3440} \\
SDXL-Turbo~\cite{podell2023sdxl} 
& \textbf{0.7647} & \textbf{0.7413} 
& \underline{0.2410} & 0.2432 
& 0.2380 & 0.2205 
& 0.1463 & 0.1232
& 0.3394 & 0.3797 
& -- & -- 
& -- & -- \\
GPT-Image-1~\cite{openai_gptimage1} 
& 0.7308 & 0.6927 
& \textbf{0.2608} & \textbf{0.2625}
& 0.1080 & \underline{0.0853} 
& 0.0680 & \underline{0.0451}
& 0.9215 & 0.9585 
& -- & -- 
& -- & -- \\
\Xhline{1.2pt} 
\end{tabular}
}
\end{table*}

\subsection{Main Results} 
\label{subsec:main_results}

We conducted a comprehensive comparison of our RMLer with existing methods: BASS~\cite{li2024tp2o}, ConceptLab~\cite{Richardson2024conceptlab}, SDXL-Turbo~\cite{podell2023sdxl}, and GPT-Image-1~\cite{openai_gptimage1}.

\textbf{Qualitative Results.}
Figs.~\ref{fig:res1} and~\ref{fig:res2} present qualitative comparisons between our method and several baselines across both ImageNet-200 and Cangjie. These examples reflect the key challenges we highlighted earlier: \textit{Conceptual Imbalance}, \textit{Superficial Combination}, and \textit{Juxtaposition Generation}. As observed, methods like BASS, ConceptLab, and SDXL-Turbo often exhibit strong bias toward one of the source concepts, resulting in imbalanced or unintegrated outputs. In contrast, our approach consistently produces semantically balanced results that preserve salient features from both inputs. Additionally, GPT-Image-1 frequently suffers from superficial fusion. For example, in the \textit{owl \& snail} case in Fig.~\ref{fig:res1}, it merely overlays a snail shell onto the owl’s back, rather than synthesizing a cohesive hybrid entity. Our method, by comparison, generates a more seamless and conceptually blended composition that better reflects the intent of fusion.

\begin{figure*}[t]
    \centering
    \begin{minipage}{.33\linewidth}
        \centering
        \includegraphics[width=0.97\linewidth]{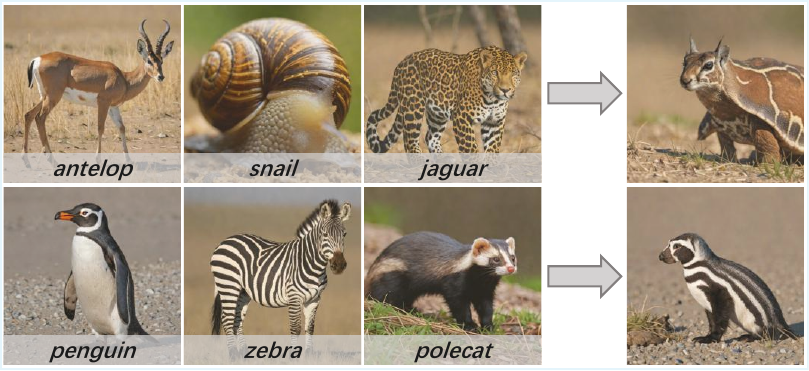} 
        \caption{Generalizations using three categroies.} 
        \label{fig:three_concepts}
    \end{minipage}
    \begin{minipage}{.27\linewidth}
        \centering
        \includegraphics[width=0.97\linewidth]{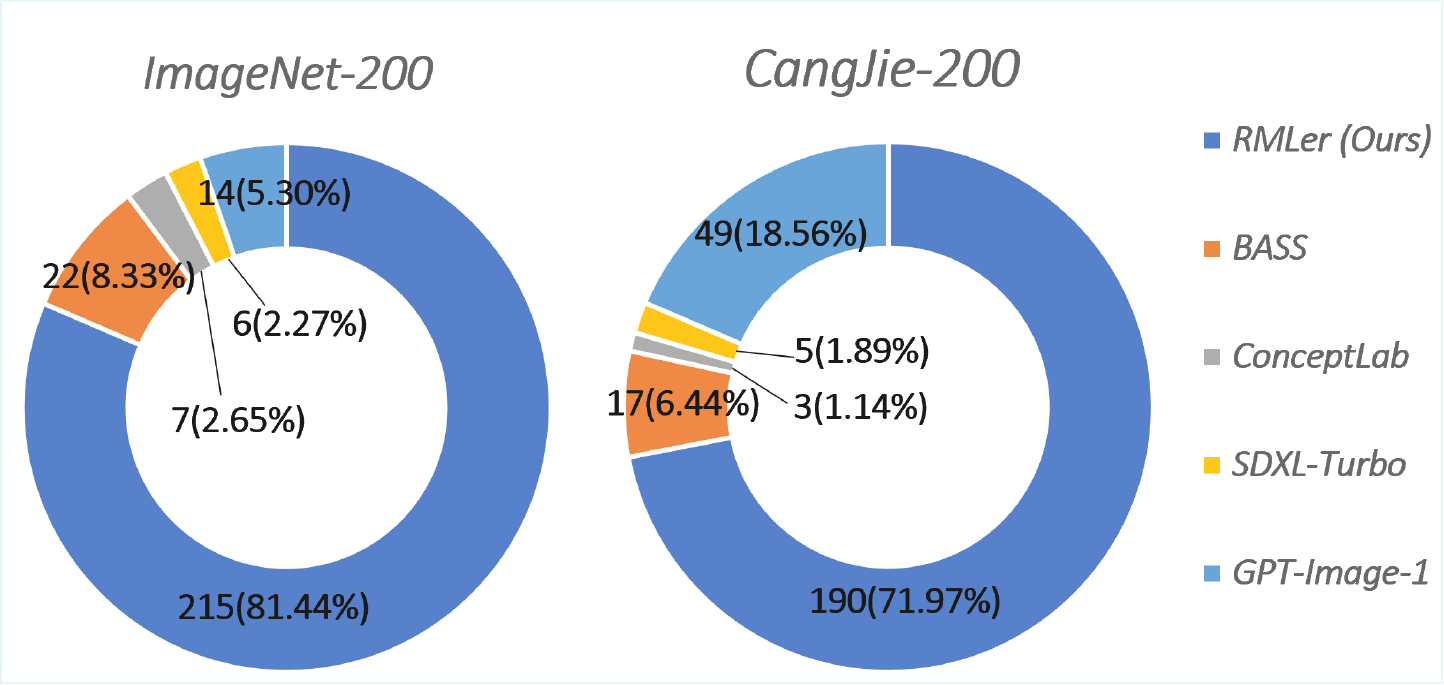} 
        \caption{User study on the \textbf{ImageNet-200} and \textbf{CangJie-200}.} 
        \label{fig:userstudy}
    \end{minipage}
    \begin{minipage}{.37\linewidth}
        \centering
        \includegraphics[width=0.97\linewidth]{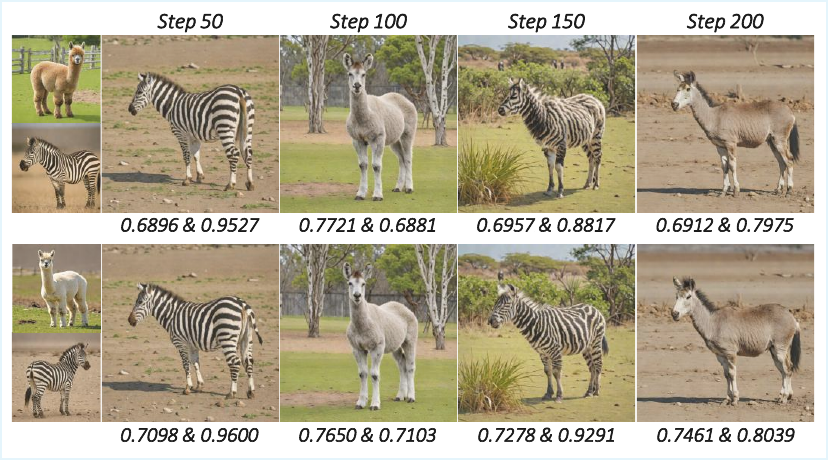} 
        \caption{Ablation study of different examplers.} 
        \label{fig:abo1}
    \end{minipage}
\end{figure*}

\begin{figure*}[t]
    \centering
    \begin{minipage}{.25\linewidth}
        \centering
        \includegraphics[width=0.97\linewidth]{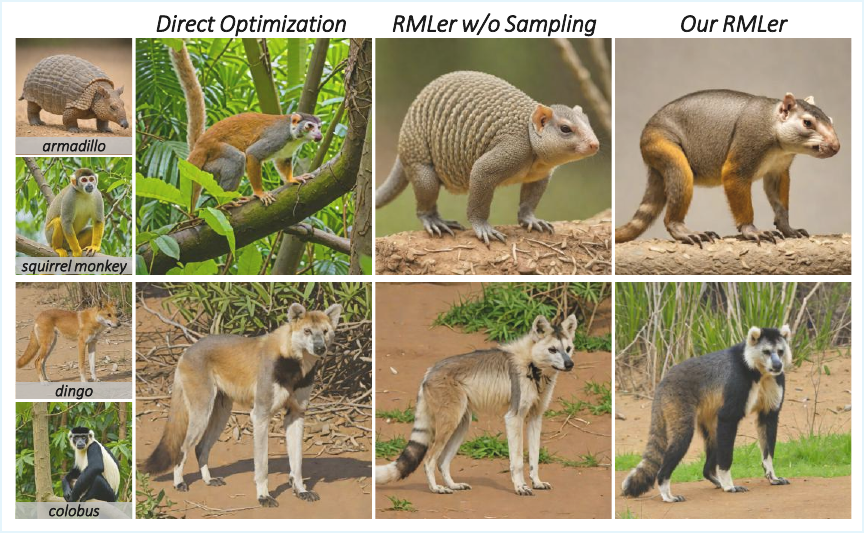} 
        \caption{Ablation study of our RMLer.} 
        \label{fig:abo2}
    \end{minipage}
    \begin{minipage}{.73\linewidth}
        \centering
    \includegraphics[width=0.97\textwidth]{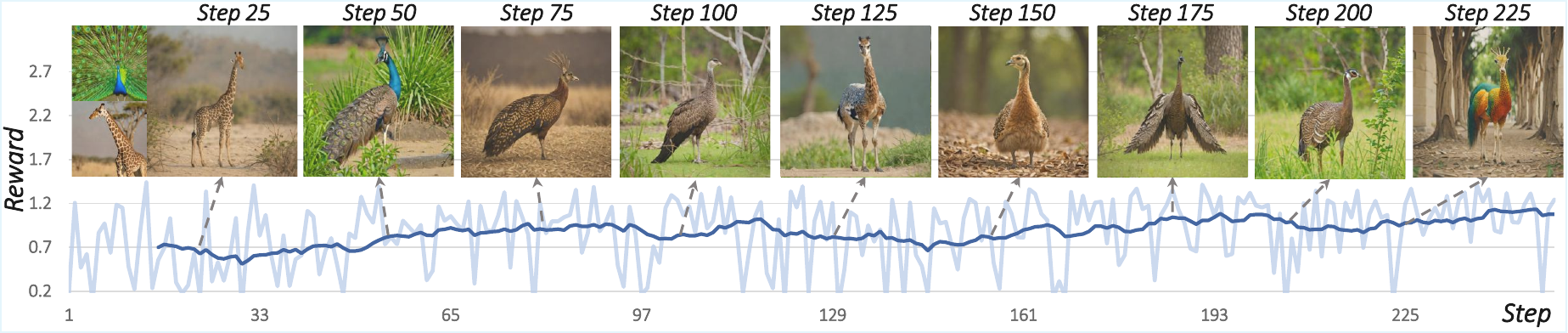} 
    \caption{Reward curve of our RMLer on the concept pair \textit{giraffe} and \textit{peacock}. } 
    \label{fig:train}
    \end{minipage}
\end{figure*}

In practice, manually conceptualizing prompts that effectively fuse two categories proves challenging. To address this, we employ GPT-4o to produce complex prompts based on RMLer-generated images. Our evaluation reveals mixed success rates when implementing these prompts with both SDXL-Turbo and GPT-Image-1. These results highlight the inherent difficulty of the C3F task, even when using state-of-the-art models like GPT-Image-1. Moreover, Fig. \ref{fig:three_concepts} demonstrates that our method can handle more than two categories.

\textbf{Quantitative Results.}
Table~\ref{tab:quant_cmp_splitcol_extended} reports quantitative comparisons across both ImageNet-200 and CangJie-200 benchmarks. Our method achieves state-of-the-art performance on key metrics, including \textit{Balance}, \textit{Reward}, \textit{HPSv2}, and \textit{VQAScore}, consistently outperforming the related approaches. These results demonstrate that RMLer not only generates more semantically balanced fusion images but also produces outputs with stronger visual appeal and better alignment with human preferences. Our Avg. Sim (I$\rightarrow$I) scores are slightly lower than SDXL-Turbo’s, this is because SDXL-Turbo generates high-quality composites of both categories (see Fig. \ref{fig:res1}) rather than true concept fusion. Similarly, our Avg. Sim (I$\rightarrow$T) scores are lower than GPT-Image-1’s, as GPT-Image-1 generates object-spliced partial semantic information instead of genuine concept fusion. Due to this discrepancy, we do not use image-text similarity as a reward signal. Additionally, we exclude HPSv2 and VQAScore for GPT-Image-1 and SDXL-Turbo, as these metrics assess image-text alignment using the input prompt. Since both models generate images directly from the evaluation prompt, their scores would be artificially inflated and incomparable.





\textbf{User Study.}
To evaluate perceptual preference for fused images, we conducted a user study on both ImageNet-200 and CangJie-200, comparing our RMLer with four existing methods: BASS, ConceptLab, SDXL-Turbo, and GPT-Image-1. A total of 66 participants cast 528 votes by selecting the most harmonious fusion result per concept pair. As shown in Fig. \ref{fig:userstudy}, RMLer received the highest preference on both benchmarks, achieving 81.44\% on ImageNet-200 and 71.97\% on CangJie, significantly outperforming all baselines. GPT-Image-1 ranked second on CangJie with 18.56\%, while other methods received substantially lower preference rates.


\subsection{Ablation Study and Parameter Analysis}

\textbf{Ablation Study.}  
To compute similarity-based rewards during training, we use pre-generated exemplar images for each concept. To evaluate whether exemplar selection affects fusion quality, we analyze variations in exemplar sets. In Fig.~\ref{fig:abo1}, the visual appearance of generated results remains largely consistent, demonstrating robustness to exemplar choice. However, minor fluctuations in quantitative metrics (e.g., similarity scores) can occur. \textbf{\textit{See Appx.~B for further analysis.}}

We further conduct ablation studies to evaluate RMLer’s core components: the RL-trained policy and stochastic sampling. In Fig.~\ref{fig:abo2}, disabling either component (e.g., substituting reward-guided optimization for the RL policy or using deterministic sampling) degrades fusion quality and introduces semantic imbalance. This confirms that adaptive policy learning and controlled stochasticity are both critical for coherent, balanced concept fusion. \textbf{\textit{See Appx.~C for extended analysis and failure cases.}}




\textbf{Parameter Analysis.}
\label{subsec:parameter_analysis}
Key hyperparameters in our RMLer include the balance factor $\alpha$ in the reward function (Eq.~\eqref{eq:reward_function_final}), the training steps and the thresholds $\tau_{\text{presence}}$ and $\tau_{\text{balance}}$ in the candidate image selection process.

\textit{Reward Balance $\alpha$.} We selected 10 diverse concept pairs from our ImageNet-200 benchmark to evaluated the reward balance factor $\alpha \in \{0, 1, 3, 5, 7\}$, training a separate RMLer agent for each configuration. For each $\alpha$, we generated 100 fused images (10 per pair) and assessed them using HPSv2 and VQAScore. As Table~\ref{tab:alpha_ablation} shows, $\alpha=5$ achieves the optimal balance between fusion quality and concept preservation.

\textit{Training Steps.} Fig. \ref{fig:train} illustrates RMLer's training dynamics for the  \textit{giraffe}-\textit{peacock} concept pair. The reward curve demonstrates consistent improvement across training iterations, reflecting progressively better concept fusion quality. This optimization process enables the framework to ultimately generate high-fidelity fused outputs.

\textit{Selection Thresholds $\tau_{\text{presence}}$ and $\tau_{\text{balance}}$.} For filtering fused results, we empirically set $\tau_{\text{presence}}=0.63$ to ensure both source concepts are sufficiently present, and $\tau_{\text{balance}}=0.05$ to encourage highly symmetric fusion. \textbf{\textit{Please refer to Appx.~D for further sensitivity analysis.}}

\begin{table}[t]
\centering
\small
\caption{Parameter analysis of the reward balance factor \(\alpha\).}
\label{tab:alpha_ablation}
\setlength{\tabcolsep}{3pt}
\renewcommand{\arraystretch}{1.1}
\scalebox{1.0}{
\begin{tabular}{c|ccccc}
\Xhline{1.2pt} 
\(\alpha\) &  0   &   1  &  3  &   \textbf{ 5 } &   7 \\
\hline  
HPSv2 \(\uparrow\) & \textbf{0.2753} & 0.2747 &  0.2741  &  0.2748 &  0.2736 \\
VQAScore \(\uparrow\) & 0.1773 &  0.1751 & 0.1836 & \textbf{0.2148} & 0.1994 \\
\Xhline{1.2pt} 
\end{tabular} }
\end{table}




\section{Conclusion}
\label{sec:conclusion}
In this work, we proposed RMLer, the first reinforcement learning (RL) framework for concept fusion in text-to-image synthesis. Leveraging PPO, our method learns an adaptive policy to dynamically manipulate text embeddings, enabling precise control over concept fusion in diffusion models. We designed a CLIP-based viusal reward function that ensures semantically coherent and well-balanced generations, along with a novel selection mechanism to identify the most representative fused output. Extensive experiments on two diverse benchmarks show that RMLer outperforms the related fused methods by a significant margin—both in quantitative metrics and human evaluations—particularly when mixing semantically dissimilar concepts. \textbf{\textit{See Appx. E for limitations.}}

\section{Acknowledgments}
\label{sec:acknowledgements}
This work was supported by the National Natural Science Foundation of China under Grant Nos. U24A20330, 62361166670, 62502208 and the Youth Science Foundation of Jiangsu Province under Grant BK20230924.

\bibliography{aaai2026}

\newpage

The supplementary material provides additional experimental details and qualitative results to support the findings of the main paper. Specifically, it includes:

\begin{itemize}
    \item \textbf{Ablation Study}: Evaluations to assess the importance of core design components, including policy learning and sampling strategies.
    \item \textbf{Robustness Analysis}: We analyze the impact of exemplar variation and find our method consistently produces stable results, demonstrating robustness to initialization changes.
    \item \textbf{Failure Cases}: We present representative failure cases to highlight current limitations.
    \item \textbf{Parameter Analysis}: Empirical studies on key hyperparameters, including the reward balance factor $\alpha$, and insights into their impact on fusion quality.
    \item \textbf{Limitation}: An outline of known limitations and potential future improvements.
    \item \textbf{Dataset}: A full listing of the curated \textit{ImageNet-200} benchmark and a brief summary of \textit{CangJie-200} used in our evaluation.
    \item \textbf{User Study}: A breakdown of the user study protocol, voting interface, and full vote distributions across all tested methods.
    \item \textbf{More Results}: Additional visual comparisons of fused outputs across diverse concept pairs, demonstrating the generality and consistency of our approach.
\end{itemize}






\section{A. Ablation Study}
\label{sec:sup_ablation}

\textbf{Policy and Stochasticity Ablation.}
We also conduct ablations to assess the contributions of two critical components in RMLer: the reinforcement-learned policy and its stochastic sampling strategy. In one setting, we remove stochasticity by using the mean action vector instead of sampling from the learned distribution. In another setting, we eliminate the policy altogether and replace it with direct reward-guided optimization.

Both simplifications lead to a noticeable drop in fusion quality. The deterministic policy results in rigid and less diverse outputs, while the reward-only optimization tends to generate semantically imbalanced images that strongly favor one of the input concepts. As shown in Figure~\ref{fig:abo2}, these failure cases highlight that both adaptive policy learning and stochastic exploration are essential to producing visually harmonious and conceptually blended results.

\textbf{Comparison of Reinforcement Learning Algorithms.}
To evaluate the effectiveness of our reinforcement learning formulation, we compare the proposed method with two alternative optimization strategies: standard PPO and GRPO. For a fair comparison, all agents are trained under identical settings using the same reward function and exemplar sets.

\begin{figure}[t]
    \centering
    \includegraphics[width=\linewidth]{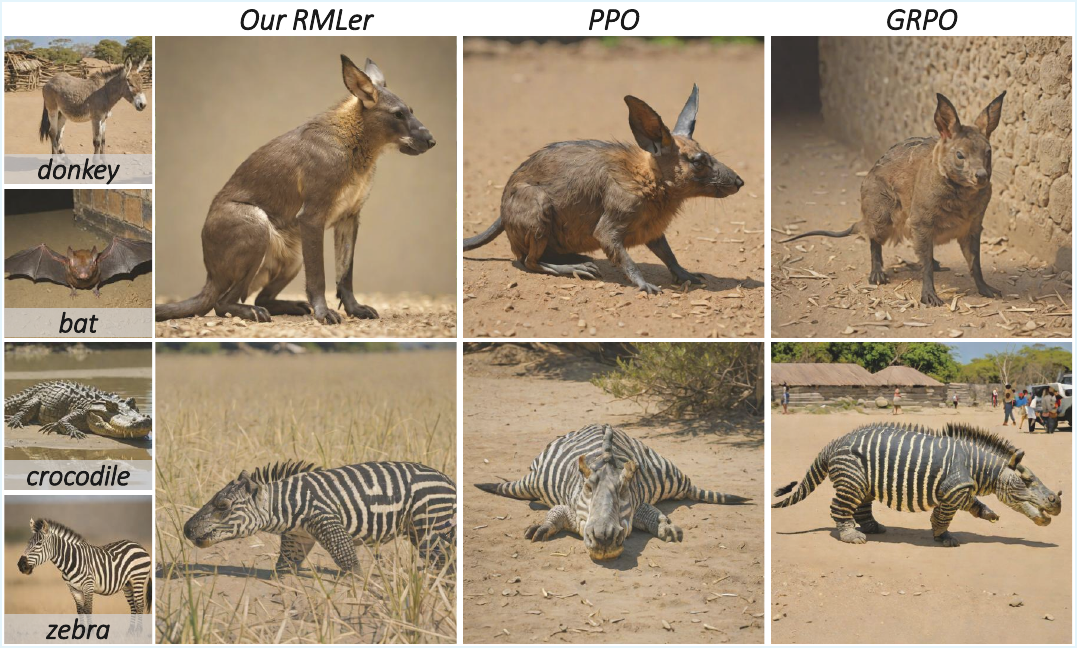}
    \caption{Ablation study comparing our full method, standard PPO, and GRPO on the same concept pair. The figure illustrates how different reinforcement learning strategies affect fusion quality, with differences in structural integration and semantic balance.}
    \label{fig:RLmethod}
\end{figure}

As shown in Figure~\ref{fig:RLmethod}, our method consistently yields higher-quality fusion results with better visual coherence and semantic balance. Quantitative evaluations, summarized in Table~\ref{tab:ppo_ablation}, further support these findings: our approach achieves superior performance across multiple metrics, indicating more stable and effective policy learning. These results validate the advantage of our tailored optimization strategy in producing precise and robust fusion across diverse concept pairs.

\textbf{Impact of the Underlying Diffusion Pipeline.}
To assess the generality of our method across different backbone generative models, we integrate our RMLer framework with four variants of the Stable Diffusion pipeline: v1.4, v1.5, v2.1, and SDXL Turbo. These models differ significantly in architecture capacity, training data, and visual styles, providing a comprehensive testbed for cross-backbone evaluation.

\begin{figure*}[t]
    \centering
    \includegraphics[width=1.0\textwidth]{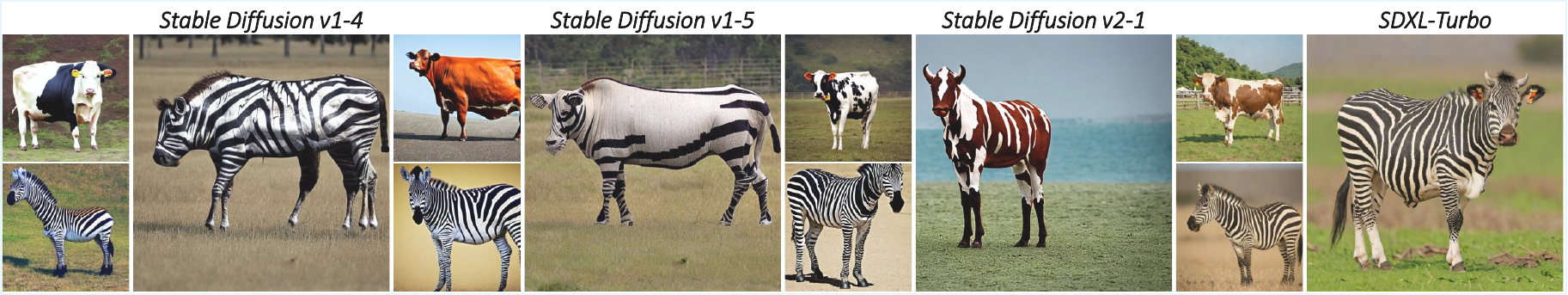}
    \caption{
    Ablation study on backbone diffusion pipelines using the concept pair \textit{Cow \& Zebra}.
    }
    \label{fig:pipeline_abo}
\end{figure*}

Figure~\ref{fig:pipeline_abo} presents fusion results under each setting. For v1.4, v1.5, and v2.1, inference is performed using 50 denoising steps, while SDXL Turbo operates with only 4 steps, offering significantly faster sampling. Despite this disparity in generation speed, our method maintains high-quality fusion across all versions. Notably, the results with SDXL Turbo not only demonstrate substantial speed improvements but also exhibit enhanced fusion quality, suggesting that more powerful backbones further amplify the effectiveness of our framework. These findings highlight the compatibility of our policy learning and reward mechanisms with a broad range of diffusion models, and underscore the modularity and efficiency of our approach.

\begin{figure*}[t]
    \centering
    \includegraphics[width=1.0\textwidth]{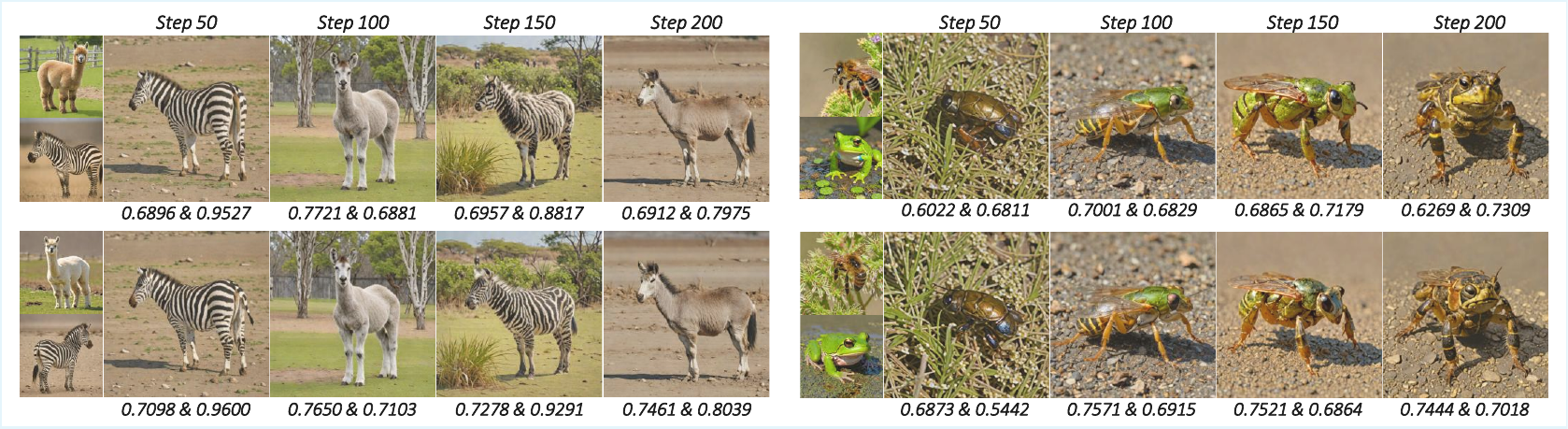}
    \caption{
    Exemplar ablation study. Each row uses a different randomly initialized exemplar set. Columns show generated images at various training steps (50 to 200). Numbers under each image denote the CLIP similarity between the fused image and the two respective exemplars (leftmost).
    }
    \label{fig:abo_sup}
\end{figure*}

\section{B. Robustness Analysis}
\label{sec:further_abo}
\textbf{Exemplar Robustness.}
During reinforcement learning, we compute similarity-based rewards using a fixed set of pre-generated exemplar images for each input concept. These exemplars act as visual anchors, guiding the learning process through CLIP-based similarity comparisons to promote semantically aligned and visually balanced fusion.

A natural question arises: \textit{how sensitive is our method to the specific choice of exemplars}? To investigate this, we conduct an ablation study by varying the random seed used for generating exemplar images, thus creating different visual references for the same concept pair. We retrain RMLer agents using these alternative exemplar sets, keeping all other training parameters fixed.

Figure~\ref{fig:abo_sup} illustrates the results. Each row corresponds to a unique exemplar set, and each column shows outputs at different training steps (50, 100, 150, 200). The two numbers below each image indicate the CLIP similarity scores between the fused result and the two source exemplars (leftmost images). These scores reflect the semantic alignment between the generated image and its reference concepts.

Despite the exemplar variation, the generated outputs remain visually consistent, indicating that our learning framework is robust to changes in exemplar initialization. While small fluctuations are observed in similarity scores, they do not meaningfully affect the training dynamics or the perceptual quality of the fusion results. This supports the stability and generality of our reward formulation under realistic exemplar variation.

\begin{figure}[t]
    \centering
    \includegraphics[width=0.47\textwidth]{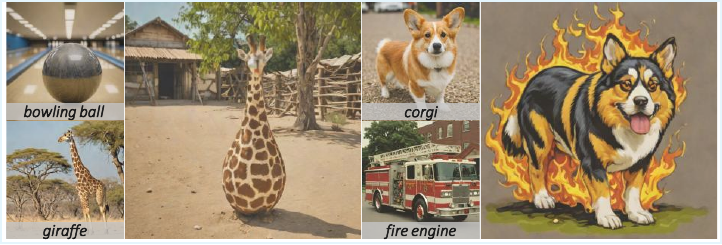}
    \caption{
    Representative failure cases of RMLer. When presented with concept pairs of extreme semantic or structural disparity, the model may produce outputs that overly favor one source concept or capture only superficial features such as color or texture.
    }
    \label{fig:failure_examples}
\end{figure}

\section{C. Failure Cases}
\label{sec:failure_cases}

Figure~\ref{fig:failure_examples} provides representative failure cases to supplement the main results. These examples illustrate scenarios where the fusion process fails to integrate features from both source concepts effectively.

A typical failure occurs when fusing objects with highly divergent semantic and structural characteristics—particularly between biological entities (e.g., animals) and rigid man-made artifacts (e.g., furniture, appliances). For example, when combining a \textit{corgi} with a \textit{coffee machine}, the model often produces results dominated by one concept, or captures only superficial traits such as color or surface texture, without meaningful structural fusion.

\begin{table}[ht]
\centering
\small
\caption{Parameter analysis of the reward balance factor \(\alpha\) in Eq.~\eqref{eq:reward_function_final}. Results are averaged over 100 samples per \(\alpha\) setting. Higher is better.}
\vspace{2pt}
\label{tab:alpha_ablation_sup}
\renewcommand{\arraystretch}{1.1}
\begin{tabular}{c|cc}
\toprule
\(\alpha\) & HPSv2 \(\uparrow\) & VQAScore \(\uparrow\) \\
\midrule
0 & \textbf{0.2753} & 0.1773 \\
1 & 0.2747 & 0.1751 \\
2 & 0.2746 & 0.1899 \\
3 & 0.2741 & 0.1836 \\
4 & 0.2729 & 0.2094 \\
\textbf{5} & 0.2748 & \textbf{0.2148} \\
6 & 0.2746 & 0.2096 \\
7 & 0.2736 & 0.1994 \\
8 & 0.2742 & 0.1781 \\
\bottomrule
\end{tabular}
\end{table}

\begin{figure*}[t]
    \centering
    \includegraphics[width=1.0\textwidth]{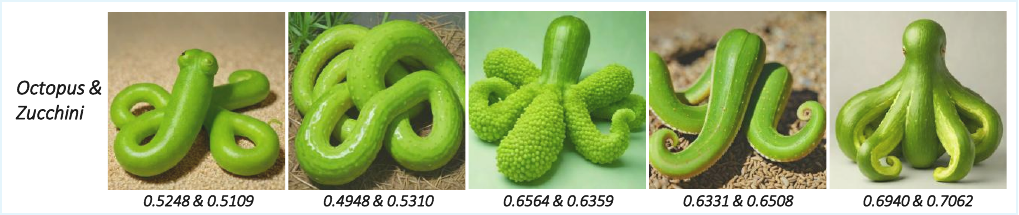}
    \caption{
    Parameter analysis on selection thresholds using the concept pair \textit{Octopus \& Zucchini}. 
    Each image is generated with different threshold settings. The two numbers below each image indicate the CLIP similarity between the generated image and the pre-generated exemplar images of the two source concepts, respectively.
    This highlights the impact of \(\tau_{\text{presence}}\) and \(\tau_{\text{balance}}\) on concept coverage and fusion symmetry.
    }
    \label{fig:thresh_analysis}
\end{figure*}

\section{D. Parameter Analysis}
\label{sec:sup_param}

\textbf{Reward Balance Factor \(\alpha\).}  
To evaluate the effect of the reward balance factor \(\alpha\) in Eq.~\eqref{eq:reward_function_final}, we conducted a controlled experiment across \(\alpha \in \{0, 1, 2, 3, 4, 5, 6, 7, 8\}\). For each setting, we selected 10 diverse concept pairs from the \textbf{ImageNet-200} benchmark and trained a separate agent for each pair, resulting in 90 independently trained models. After training, we used the final checkpoint of each model to generate 10 fused samples per pair, totaling 100 images per \(\alpha\). These samples were evaluated using \textbf{HPSv2} and \textbf{VQAScore}, with results averaged across each group. Table~\ref{tab:alpha_ablation_sup} reports the mean scores.

We observe that \(\alpha=5\) achieves the highest VQAScore (0.2148), indicating optimal semantic alignment. Although \(\alpha=0\) yields the best HPSv2 score (0.2753), it performs poorly on VQAScore, suggesting that visually appealing results may lack semantic balance. Conversely, larger values such as \(\alpha=7\) and \(\alpha=8\) reduce performance across both metrics, likely due to over-penalization of imbalance. Based on this analysis, we adopt \(\alpha=5\) as the default configuration, offering the best trade-off between semantic coverage and fusion symmetry.

\textbf{Selection Thresholds $\tau_{\text{presence}}$ and $\tau_{\text{balance}}$.}  
To ensure both semantic coverage and fusion symmetry in selected results, we apply a two-stage filtering strategy based on thresholds $\tau_{\text{presence}}$ and $\tau_{\text{balance}}$ (see Sec.~\ref{subsec:image_selection}). The presence threshold $\tau_{\text{presence}}$ ensures both source concepts are sufficiently expressed, by requiring their CLIP similarities (with pre-generated exemplars) to exceed a minimum value. The balance threshold $\tau_{\text{balance}}$ limits the absolute difference between these two similarities, promoting symmetric concept fusion.

We conducted sensitivity analysis to determine optimal values. For $\tau_{\text{presence}}$, values below 0.60 allowed noisy or semantically sparse samples, while values above 0.65 were overly restrictive. We selected $\tau_{\text{presence}} = 0.63$ as a balance between precision and coverage. Similarly, we varied $\tau_{\text{balance}}$ from 0.01 to 0.1. Larger values admitted asymmetric fusions favoring a dominant concept, while smaller values filtered more aggressively. We found $\tau_{\text{balance}} = 0.05$ provided the best trade-off.

Figure~\ref{fig:thresh_analysis} illustrates the effect of different threshold settings using the concept pair \textit{Octopus \& Zucchini}. The two values below each image denote CLIP similarities to the two source concept exemplars, reflecting semantic presence and balance. Our chosen thresholds consistently retain high-quality, well-integrated fusions across datasets.

\begin{table*}[ht]
\centering
\small
\caption{User study vote counts and preference rates (\%) on \textbf{ImageNet-200} and \textbf{CangJie-200}.}
\label{tab:app_user_votes}
\resizebox{0.95\linewidth}{!}{
\begin{tabular}{l|ccccc}
\toprule
Dataset & RMLer (Ours) & BASS~\cite{li2024tp2o} & ConceptLab~\cite{Richardson2024conceptlab} & SDXL-Turbo~\cite{podell2023sdxl} & GPT-Image-1~\cite{openai_gptimage1} \\
\midrule
ImageNet-200 (Votes) & \textbf{215} & 22 & 7 & 6 & 14 \\
ImageNet-200 (\%)    & \textbf{81.44\%} & 8.33\% & 2.65\% & 2.27\% & 5.30\% \\
\midrule
CangJie-200 (Votes)  & \textbf{190} & 17 & 3 & 5 & 49 \\
CangJie-200 (\%)     & \textbf{71.97\%} & 6.44\% & 1.14\% & 1.89\% & 18.56\% \\
\bottomrule
\end{tabular}
}
\end{table*}

\section{E. Limitation}
\label{sec:limitation}

Despite the strong performance of RMLer in synthesizing visually balanced and semantically coherent fusions, the framework exhibits several limitations when handling inputs with large representational gaps.

Our analysis shows that extreme semantic or structural disparity—especially across concept domains such as biological versus non-biological entities—poses a challenge for the current fusion mechanism. The policy may collapse toward a dominant concept, or fail to preserve part-level correspondences, due to difficulties in aligning incompatible latent spaces and the limitations of similarity-based rewards in capturing higher-order structure.

To address these issues, future work could explore structured fusion strategies that incorporate part-aware representations, scene-level constraints, or disentangled embeddings for shape and texture. Additionally, augmenting the reward function with structural or commonsense priors, or extending training to more diverse and compositional datasets, may enhance generalization and robustness to challenging concept pairs. Inspired by large-scale model training paradigms, incorporating a supervised or heuristic-guided warm-up phase before reinforcement learning may also stabilize policy optimization and provide a stronger initialization for complex fusion tasks.

\section{F. Dataset}
\label{sec:dataset}

We evaluate our method on a benchmark of 400 diverse concept pairs, divided into two subsets. The first, \textbf{ImageNet-200}, consists of 200 manually curated pairs from the ImageNet~\cite{russakovsky2015imagenet} vocabulary. We selected semantically and visually dissimilar concepts while avoiding overlapping or hierarchical categories (e.g., ``squirrel'' vs. ``squirrel monkey''), ensuring a challenging and meaningful fusion task. The full list is shown in Table~\ref{tab:imagenet_dataset}.

The second subset, \textbf{CangJie-200}, is adapted from the \textit{CangJie} dataset introduced in CreTok~\cite{feng2024redefining}, designed to benchmark combinatorial creativity in the TP2O~\cite{li2024tp2o} setting. It comprises 200 pairs involving animals and plants (e.g., dogs, monkeys, pineapples), and complements ImageNet-200 by emphasizing symbolic and abstract fusion scenarios.

\section{G. User Study}
\label{sec:sup_user_stu}
 
\begin{figure*}[t]
    \centering
    \includegraphics[width=0.8\textwidth]{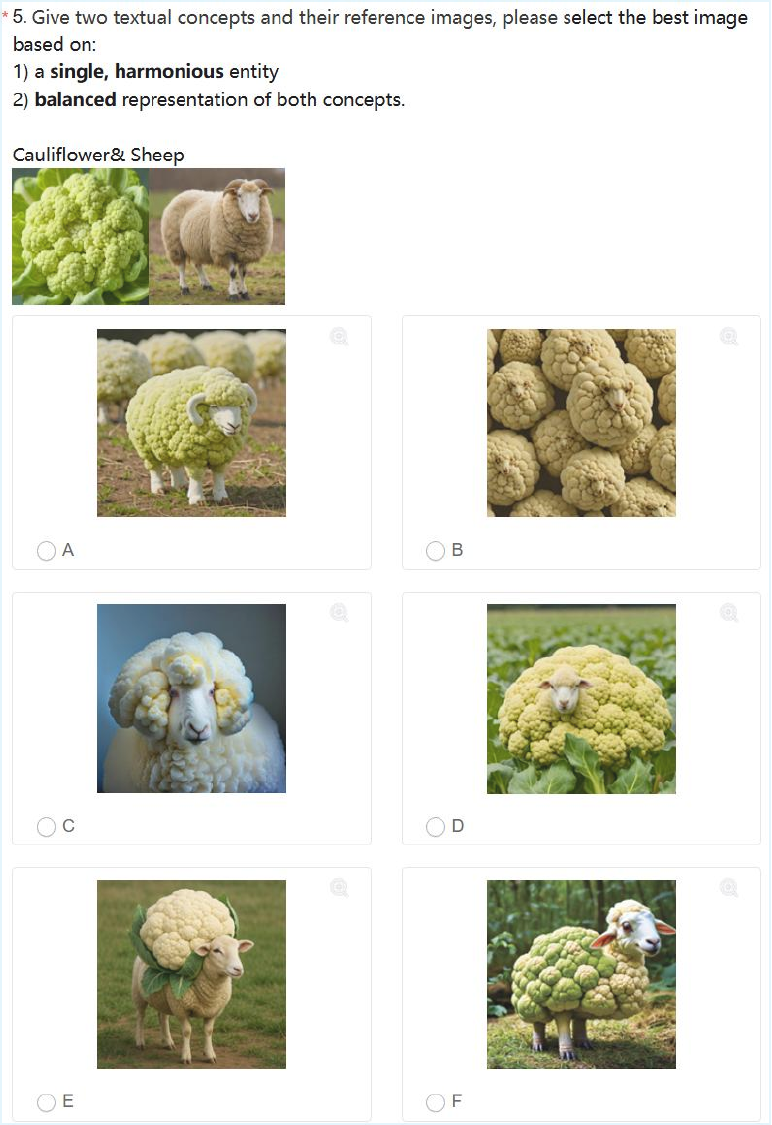}
    \vskip -0.1in
    \caption{Interface used in the user preference study. Participants viewed five anonymized outputs and selected the most balanced and creative fusion.}
    \label{fig:userstudy_interface}
\end{figure*}

To comprehensively assess the perceptual quality and creativity of the fused results, we conducted a large-scale user preference study on both \textbf{ImageNet-200} and \textbf{CangJie-200} benchmarks. The study compared our method (RMLer) with four competitive baselines: BASS~\cite{li2024tp2o}, ConceptLab~\cite{Richardson2024conceptlab}, SDXL-Turbo~\cite{podell2023sdxl}, and GPT-Image-1~\cite{openai_gptimage1}. The evaluation interface used during the study is shown in Fig.~\ref{fig:userstudy_interface}.

For each concept pair, participants were shown five images—one from each method—and asked to choose the most visually harmonious and creatively integrated fusion. In total, we collected \textbf{528 votes} from \textbf{66 participants}, each presented with randomly sampled and anonymized concept pair tasks.

\begin{figure*}[t]
    \centering
    \includegraphics[width=1\textwidth]{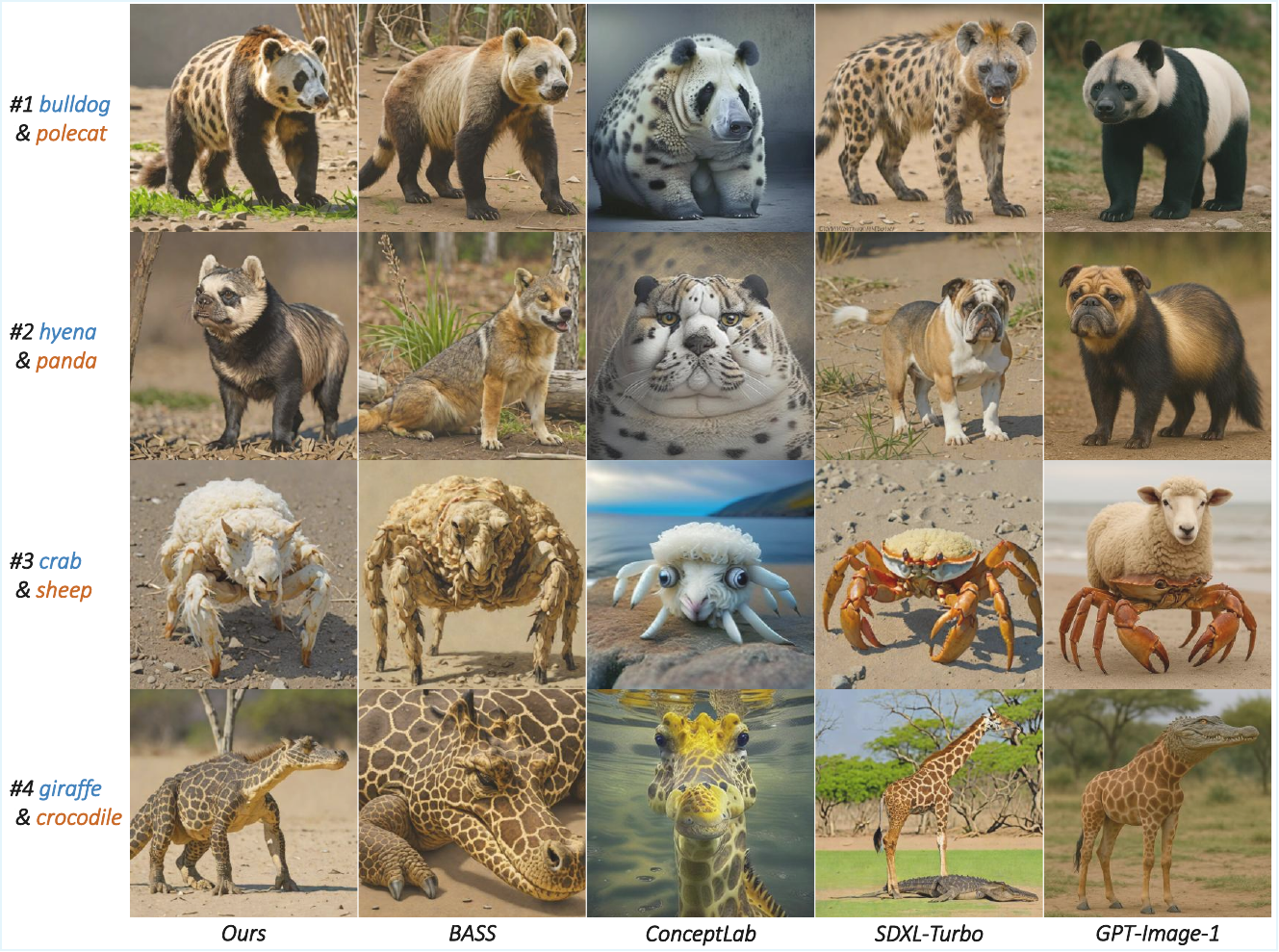}
    \caption{Representative samples from the \textbf{ImageNet-200} benchmark used in the user study. Participants selected their preferred image from each row.}
    \label{fig:app_user_examples1}
\end{figure*}

\begin{figure*}[t]
    \centering
    \includegraphics[width=1\textwidth]{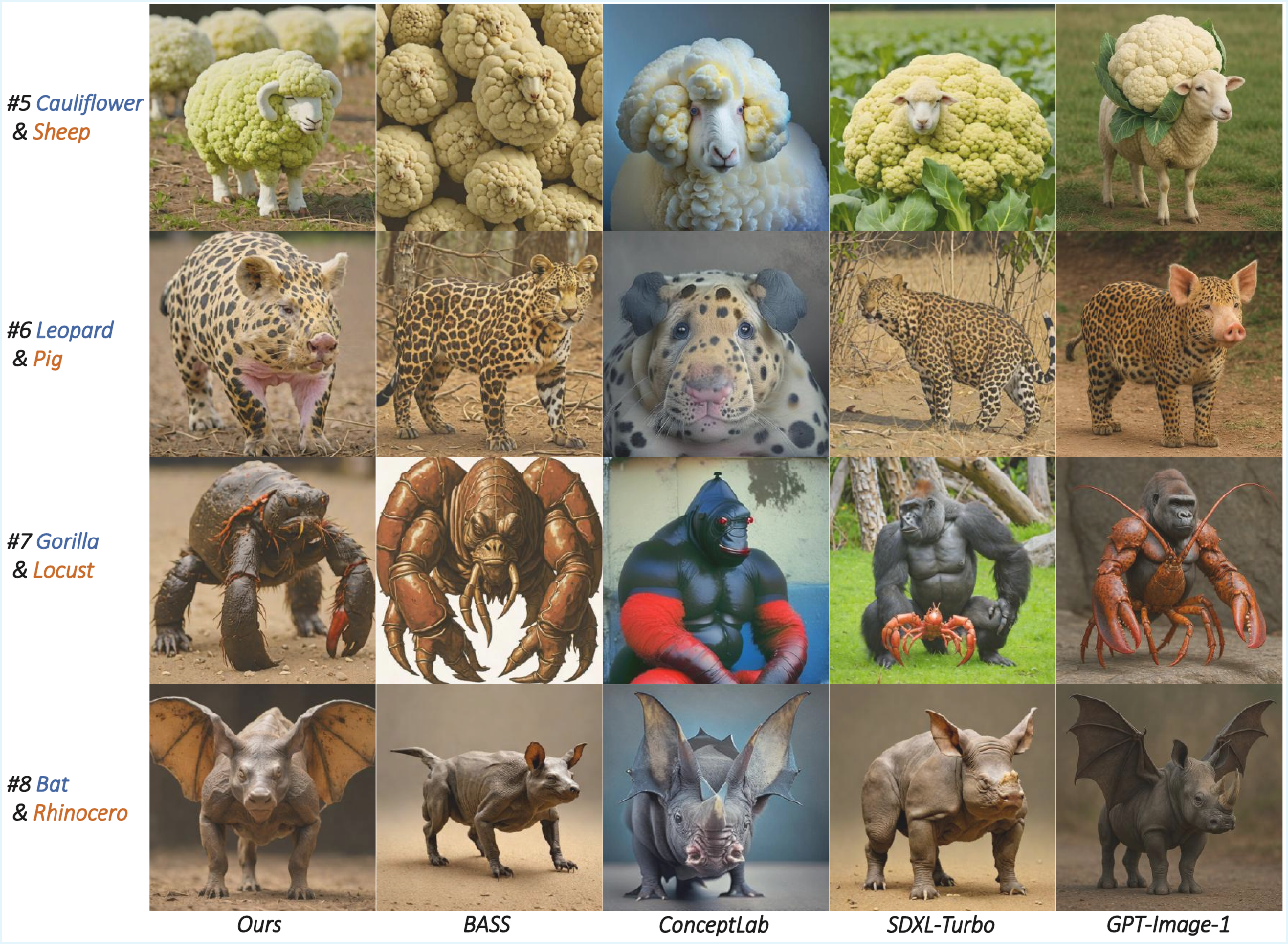}
    \caption{Representative samples from the \textbf{CangJie-200} benchmark used in the user study.}
    \label{fig:app_user_examples2}
\end{figure*}

As summarized in Table~\ref{tab:app_user_votes}, RMLer was overwhelmingly preferred across both datasets, achieving a preference rate of \textbf{81.44\%} on ImageNet-200 and \textbf{71.97\%} on CangJie-200. GPT-Image-1 ranked second on CangJie with 18.56\%, while all other baselines received significantly lower preference shares. Visual examples of the candidate comparisons shown to participants are provided in Fig.~\ref{fig:app_user_examples1} and Fig.~\ref{fig:app_user_examples2}.

\section{H. More Results}
\label{sec:sup_more}

\begin{figure*}[t]
    \centering
    \includegraphics[width=1\textwidth]{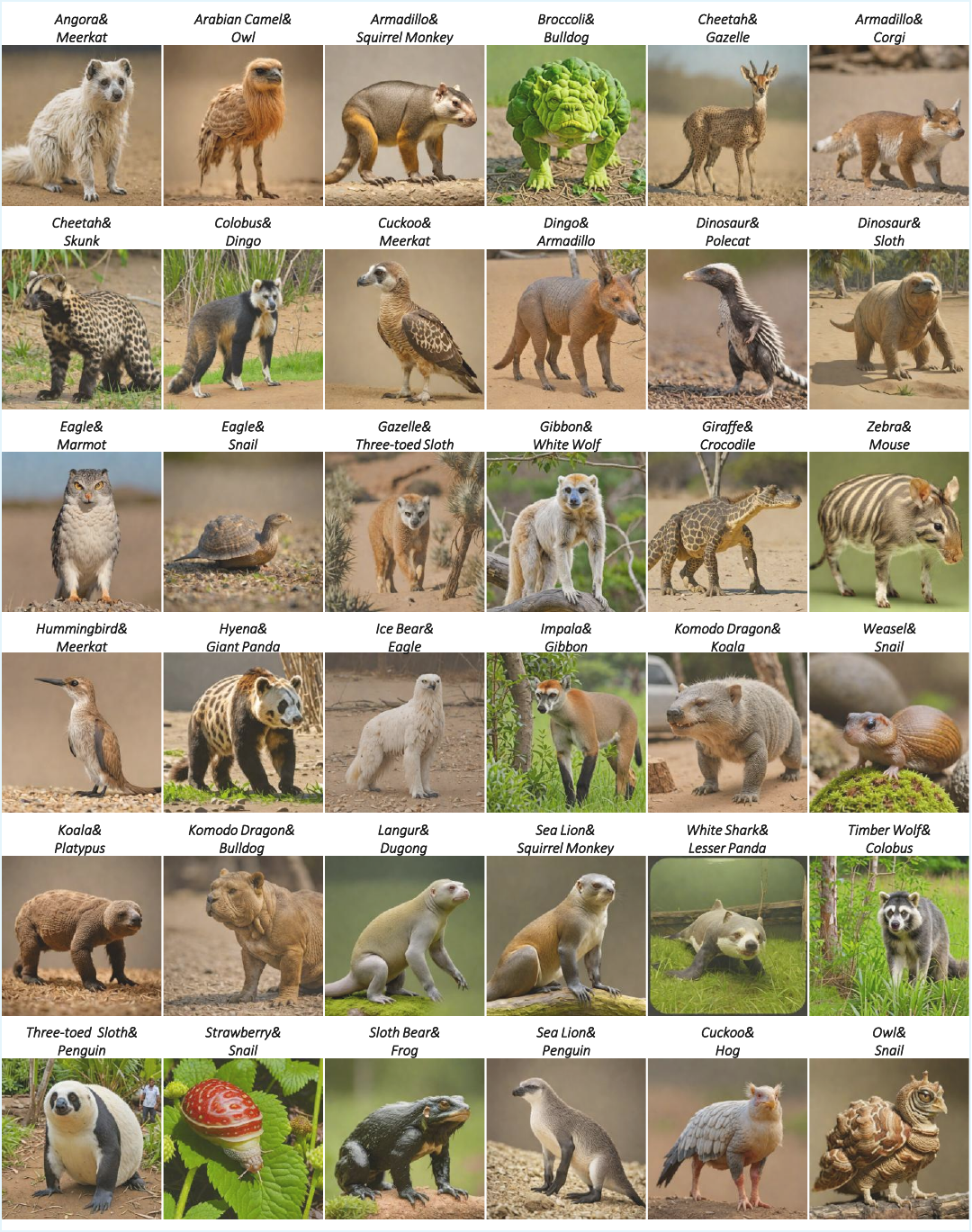}
    \caption{
    Additional fusion results generated by RMLer on concept pairs from the \textbf{ImageNet-200} benchmark. These samples illustrate the model’s ability to fuse semantically diverse and structurally rich categories with high visual realism and compositional fidelity.
    }
    \label{fig:sup_more1}
\end{figure*}

\begin{figure*}[t]
    \centering
    \includegraphics[width=1\textwidth]{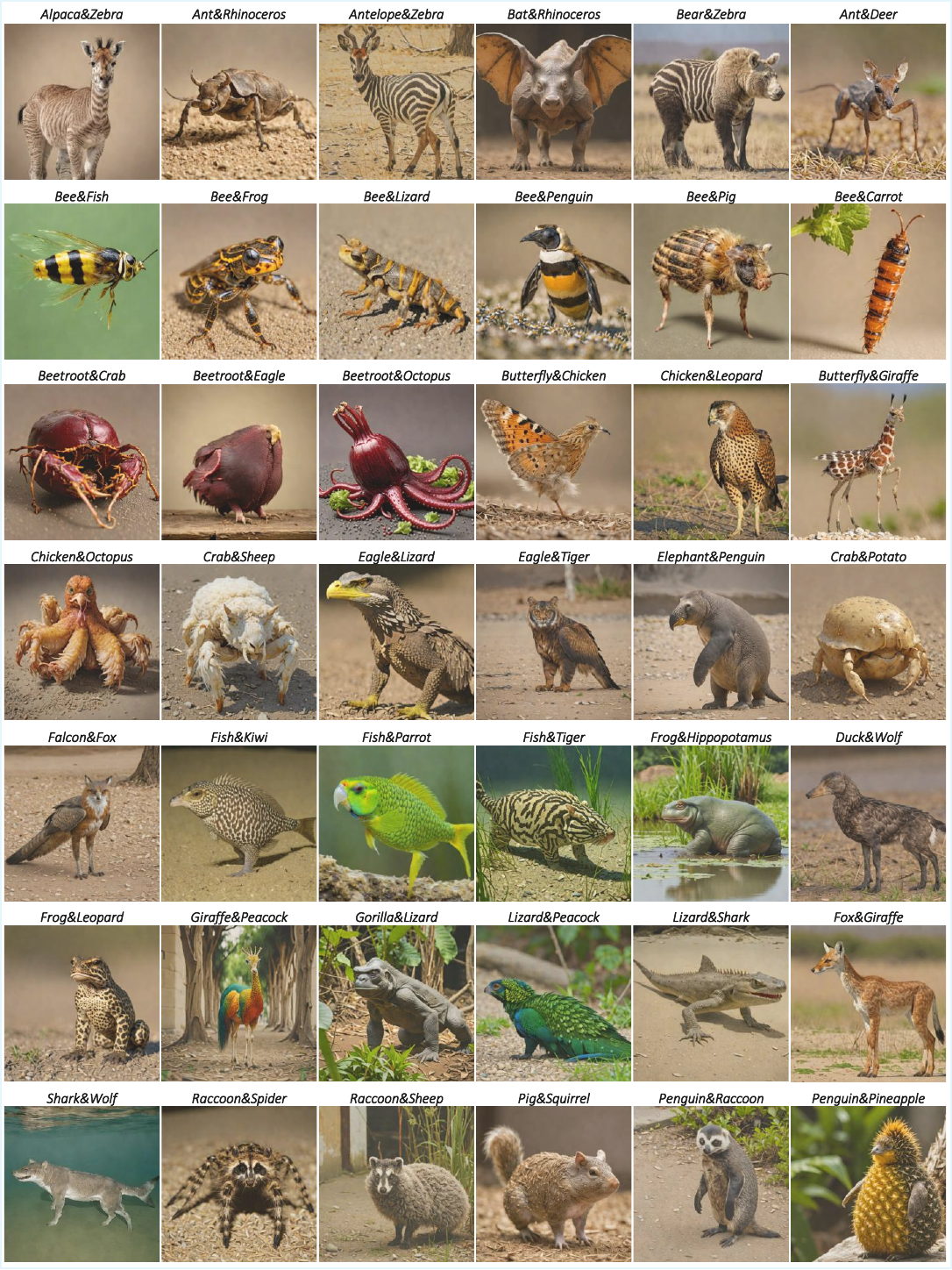}
    \caption{
    Additional fusion results generated by RMLer on concept pairs from the \textbf{CangJie-200} benchmark. These samples highlight the model’s ability to generate coherent, semantically grounded, and visually creative fusions across abstract and cross-domain concepts.
    }
    \label{fig:sup_more2}
\end{figure*}

\begin{figure*}[t]
    \centering
    \includegraphics[width=1\textwidth]{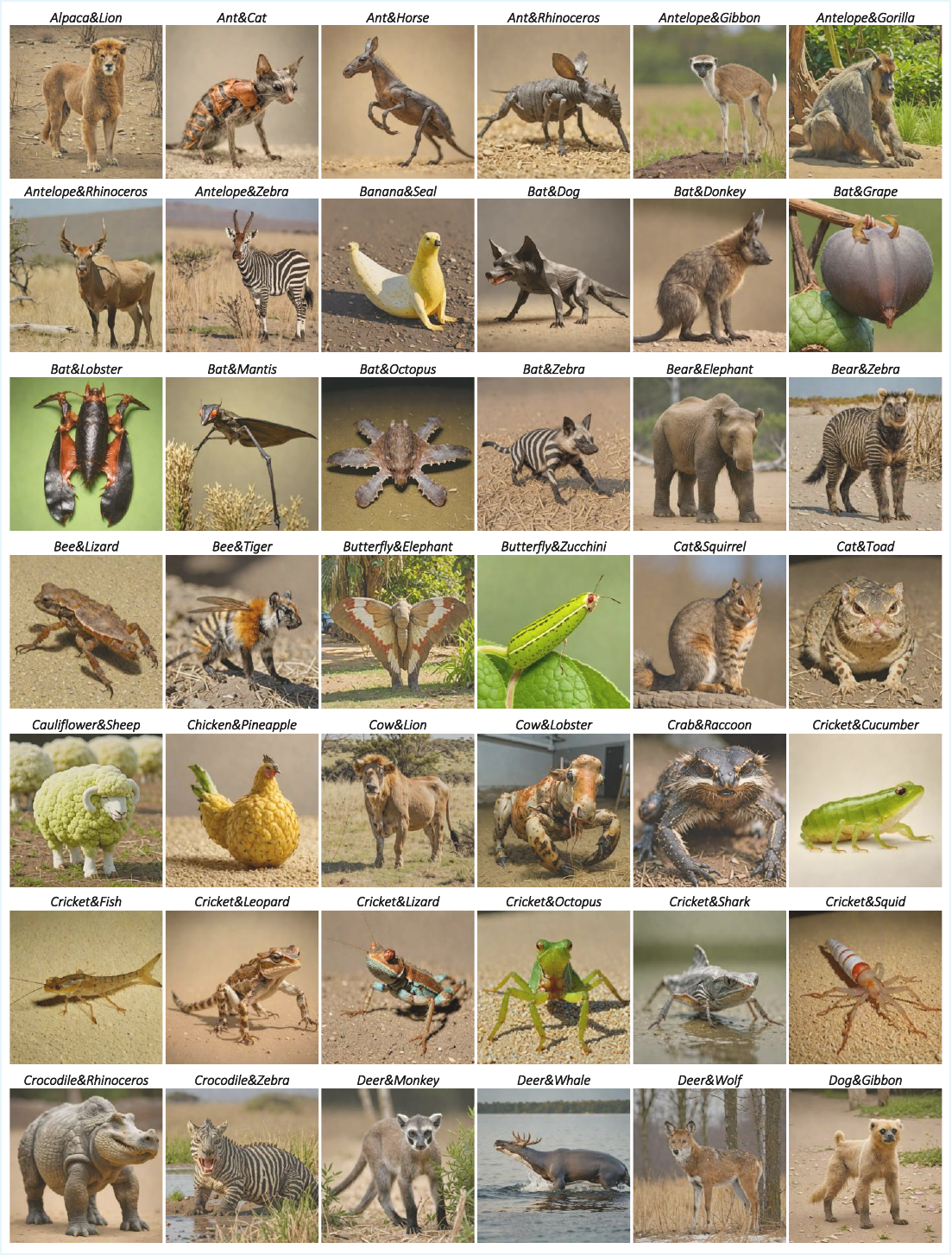}
    \caption{
    Additional fusion results generated by RMLer on concept pairs from the \textbf{CangJie-200} benchmark. These samples highlight the model’s ability to generate coherent, semantically grounded, and visually creative fusions across abstract and cross-domain concepts.
    }
    \label{fig:sup_more3}
\end{figure*}

\begin{figure*}[t]
    \centering
    \includegraphics[width=1\textwidth]{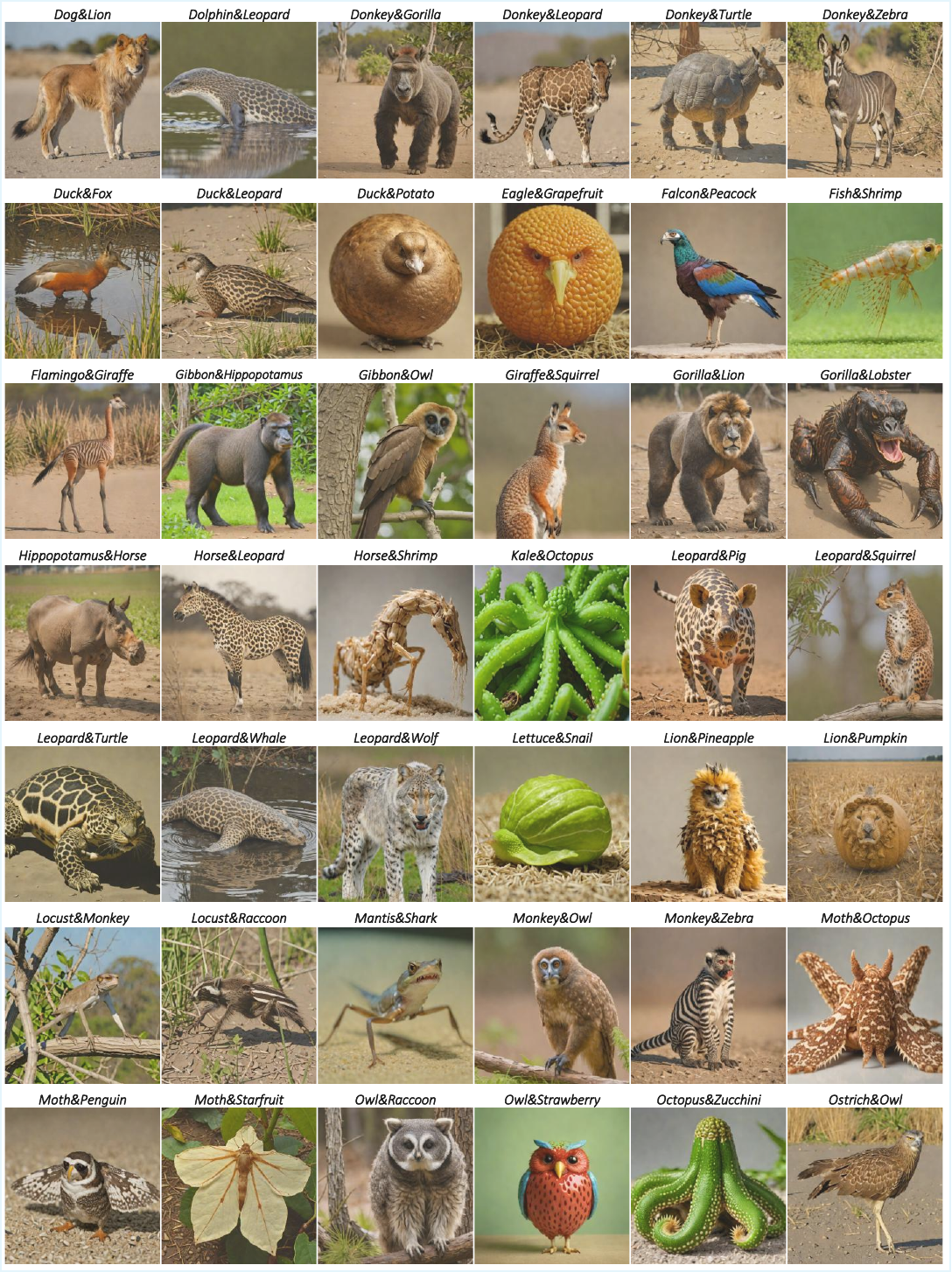}
    \caption{
    Additional fusion results generated by RMLer on concept pairs from the \textbf{CangJie-200} benchmark. These samples highlight the model’s ability to generate coherent, semantically grounded, and visually creative fusions across abstract and cross-domain concepts.
    }
    \label{fig:sup_more4}
\end{figure*}

In this section, we present additional generation results produced by our RMLer framework to further demonstrate its versatility and effectiveness in fusing diverse textual concepts. Figures~\ref{fig:sup_more1}–\ref{fig:sup_more4} showcase a wide array of fused outputs from our benchmark.

Figure~\ref{fig:sup_more1} presents results from the \textbf{ImageNet-200} subset, highlighting fusion across semantically and visually diverse concept pairs. Figures~\ref{fig:sup_more2}, \ref{fig:sup_more3}, and \ref{fig:sup_more4} show results from the \textbf{CangJie-200} subset, focusing on more abstract and compositional pairs inspired by animal-plant combinations in the TP2O setting.

These examples illustrate RMLer's ability to handle challenging fusion scenarios, particularly when the source concepts exhibit high semantic disparity. In many cases, RMLer effectively preserves salient attributes from both source concepts—such as texture, shape, or structure—while achieving coherent and visually realistic integration. The generated results reflect strong conceptual blending, structural consistency, and creative fidelity.

To further assess the generalization capability of RMLer, we also explore stylized generation during inference. The model is trained using default textual prompts, but at test time, we append additional descriptive modifiers representing different visual aesthetics. The styles include:

\textit{
``anime-style illustration, clean lineart, flat shading, soft pastel colors, simple background'';  
``cyberpunk aesthetic, neon lighting, high contrast colors, glowing elements, futuristic cityscape background, gritty texture, moody atmosphere'';  
``Van Gogh style painting, expressive brushstrokes, swirling textures, vivid and saturated colors, thick impasto oil paint, impressionist background'';  
``cubism, abstract forms, geometric distortions, flat perspective, bold outlines''.
}

\begin{figure*}[t]
    \centering
    \includegraphics[width=1\textwidth]{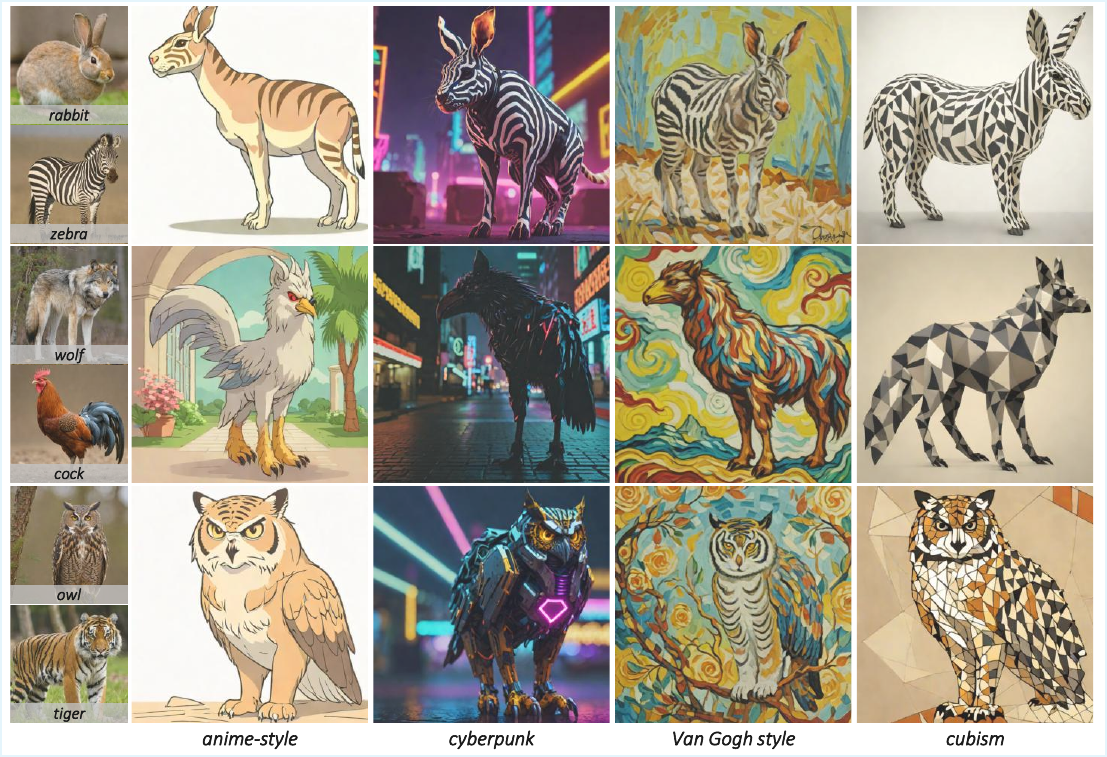}
    \caption{
    Stylized visualization of fused concept pairs across different visual aesthetics. Each row shows hybrid animals synthesized from two source species (leftmost), rendered in four artistic styles: \textbf{anime-style}, \textbf{cyberpunk}, \textbf{Van Gogh style}, and \textbf{cubism}. These results are generated using our default fusion model with style-specific prompts applied at inference time, without any model finetuning or cherry-picking. The results demonstrate RMLer’s compositional robustness under stylistic domain shifts.
    }
    \label{fig:style}
\end{figure*}

Figure~\ref{fig:style} shows stylized fusion results across these domains. All images are generated using a fixed trained policy without cherry-picking or manual post-processing. These results demonstrate that RMLer maintains compositional fidelity and fusion robustness under significant stylistic variation.

\begin{table*}[t]
    \centering
    \setlength{\tabcolsep}{1.5 mm}
    \caption{Details of ImageNet-200.}
    \resizebox{\textwidth}{!}{%
        \begin{tabular}{@{}llll@{}}
    \bottomrule[1pt]
(African Hunting Dog, Indian Elephant) & (American Black Bear, Indian Elephant) & (Angora, Indian Elephant) & (Arabian Camel, Indian Elephant) \\
(Arctic Fox, Indian Elephant) & (Armadillo, Indian Elephant) & (Black-Footed Ferret, Indian Elephant) & (American Black Bear, Macaque) \\
(Angora, Meerkat) & (Arabian Camel, Ostrich) & (Arabian Camel, Owl) & (Arctic Fox, Sea Lion) \\
(Arctic Fox, Turtle) & (Armadillo, Corgi) & (Armadillo, Marmot) & (Armadillo, Squirrel Monkey) \\
(Black-Footed Ferret, Brown Bear) & (Black-Footed Ferret, Tusker) & (Broccoli, Bulldog) & (Broccoli, Pineapple) \\
(Broccoli, Squirrel Monkey) & (Bulldog, Polecat) & (Bulldog, Venom) & (Cat, Orangutan) \\
(Cat, Three-Toed Sloth) & (Cat, Wallaby) & (Cauliflower, Hummingbird) & (Cauliflower, Octopus) \\
(Cheetah, Gazelle) & (Cheetah, Macaque) & (Cheetah, Proboscis Monkey) & (Cheetah, Skunk) \\
(Chimpanzee, Broccoli) & (Chimpanzee, Brown Bear) & (Chimpanzee, Llama) & (Cock, Hamster) \\
(Colobus, Dingo) & (Colobus, Red Wolf) & (Colobus, Turtle) & (Corgi, Kangaroo) \\
(Corgi, Porcupine) & (Cougar, Hummingbird) & (Cougar, Tusker) & (Cougar, Wood Rabbit) \\
(Coyote, Cauliflower) & (Coyote, Sea Lion) & (Crocodile, Lizard) & (Crocodile, Venom) \\
(Cuckoo, Hog) & (Cuckoo, Meerkat) & (Cuckoo, Sloth Bear) & (Dingo, Armadillo) \\
(Dinosaur, Polecat) & (Dinosaur, Sloth) & (Dugong, Komodo Dragon) & (Dugong, Siamang) \\
(Eagle, Marmot) & (Eagle, Meerkat) & (Eagle, Snail) & (Eagle, Venom) \\
(Elephant, Jaguar) & (Elephant, Snail) & (Elephant, Venom) & (Fish, Three-Toed Sloth) \\
(Fish, Venom) & (Fox Squirrel, Venom) & (Frog, Timber Wolf) & (Frog, Venom) \\
(Gazelle, Siamang) & (Gazelle, Snow Leopard) & (Gazelle, Three-Toed Sloth) & (Gibbon, Coyote) \\
(Gibbon, Meerkat) & (Gibbon, Three-Toed Sloth) & (Gibbon, Weasel) & (Gibbon, White Wolf) \\
(Giraffe, American Black Bear) & (Giraffe, Crocodile) & (Giraffe, Dinosaur) & (Giraffe, Tiger) \\
(Hartebeest, Indian Elephant) & (Hartebeest, Crocodile) & (Hartebeest, Impala) & (Hippopotamus, Dugong) \\
(Hippopotamus, White Wolf) & (Hog, Hyena) & (Hog, Macaque) & (Hog, Venom) \\
(Howler Monkey, Lesser Panda) & (Howler Monkey, Orangutan) & (Hummingbird, Hyena) & (Hummingbird, Impala) \\
(Hummingbird, Meerkat) & (Hummingbird, Owl) & (Hyena, Giant Panda) & (Hyena, Orangutan) \\
(Ice Bear, Eagle) & (Ice Bear, Hyena) & (Ice Bear, Lizard) & (Ice Bear, Squirrel) \\
(Ice Bear, Wild Boar) & (Impala, Gibbon) & (Impala, Three-Toed Sloth) & (Indri, Cat) \\
(Indri, Proboscis Monkey) & (Indri, Lizard) & (Kangaroo, Howler Monkey) & (Kangaroo, Venom) \\
(Killer Whale, Gibbon) & (Killer Whale, Owl) & (Kit Fox, Elephant) & (Kit Fox, Timber Wolf) \\
(Koala, Meerkat) & (Koala, Platypus) & (Komodo Dragon, Bulldog) & (Komodo Dragon, Koala) \\
(Langur, African Hunting Dog) & (Langur, Dugong) & (Langur, Elephant) & (Langur, Penguin) \\
(Lesser Panda, Hummingbird) & (Lesser Panda, Ibex) & (Lion, Hartebeest) & (Lion, Venom) \\
(Lizard, American Black Bear) & (Lizard, Giant Panda) & (Lizard, Eagle) & (Lizard, Frog) \\
(Llama, Brown Bear) & (Llama, Platypus) & (Macaque, Gazelle) & (Marmot, Platypus) \\
(Meerkat, American Black Bear) & (Meerkat, Lizard) & (Mouse, Owl) & (Mouse, Sloth Bear) \\
(Octopus, Rabbit) & (Octopus, Three-Toed Sloth) & (Octopus, Venom) & (Orangutan, Ibex) \\
(Orangutan, Platypus) & (Ostrich, Timber Wolf) & (Owl, Snail) & (Ox, Crocodile) \\
(Ox, Skunk) & (Penguin, Hog) & (Penguin, Venom) & (Platypus, White Shark) \\
(Polecat, Platypus) & (Polecat, Wombat) & (Porcupine, Water Buffalo) & (Proboscis Monkey, White Shark) \\
(Rabbit, Brown Bear) & (Rabbit, Indri) & (Sea Lion, Chimpanzee) & (Sea Lion, Penguin) \\
(Sea Lion, Squirrel Monkey) & (Siamang, Dingo) & (Siamang, Skunk) & (Skunk, Kit Fox) \\
(Sloth Bear, Frog) & (Sloth Bear, Venom) & (Sloth, Snow Leopard) & (Snail, American Black Bear) \\
(Snail, Cougar) & (Snail, Ice Bear) & (Snail, Komodo Dragon) & (Snail, Water Buffalo) \\
(Snow Leopard, Fox Squirrel) & (Squirrel, Proboscis Monkey) & (Strawberry, Hummingbird) & (Strawberry, Snail) \\
(Three-Toed Sloth, Penguin) & (Tiger, Cougar) & (Timber Wolf, Colobus) & (Timber Wolf, Gorilla) \\
(Timber Wolf, Jaguar) & (Timber Wolf, Snow Leopard) & (Timber Wolf, Cock) & (Turtle, Venom) \\
(Tusker, American Black Bear) & (Tusker, Ox) & (Venom, Rabbit) & (Wallaby, Rabbit) \\
(Wallaby, Snail) & (Warthog, Cock) & (Warthog, Elephant) & (Weasel, Armadillo) \\
(Weasel, Kit Fox) & (Weasel, Snail) & (White Shark, Lesser Panda) & (White Shark, Llama) \\
(White Wolf, Arctic Fox) & (White Wolf, Proboscis Monkey) & (Wild Boar, Killer Whale) & (Wild Boar, Octopus) \\
(Zebra, Giant Panda) & (Zebra, Mouse) & (Zebra, Orangutan) & (Zebra, Squirrel) \\
    \bottomrule[1pt]
    \end{tabular}%
    }
    \label{tab:imagenet_dataset}
\end{table*}

\end{document}